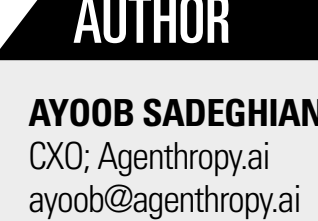

AUTHOR

**AYOOB SADEGHIANI**
CXO; Agenthropy.ai
ayoob@agenthropy.ai

RESEARCH PROCEEDING

# Generative AI Carries Non-Democratic Biases and Stereotypes: Representation of Women, Black Individuals, Age Groups, and People with Disability in AI-Generated Images across Occupations

## ABSTRACT

In this study, I investigate how generative artificial intelligence (AI) systems reproduce and reinforce societal biases, with a specific focus on the representation of women, Black individuals, age groups, and people with visible disabilities in AI-generated occupational images. I analyzed 444 images generated by Microsoft Designer, Meta AI, and Ideogram across 37 occupations and found significant disparities in representation. Women are underrepresented in senior and technology roles, Black individuals are nearly absent, and people with visible disabilities are completely absent across all categories. I also observed clear age bias, with younger individuals predominantly depicted. These patterns suggest that generative AI tools replicate, and in some cases amplify, existing workplace inequalities and stereotypes, undermining democratic values of equity and inclusion. My findings highlight the urgent need for algorithmic diversity exposure, and I recommend that AI developers and corporate users audit their tools for equity, diversity, and inclusion (EDI) risks. I argue for the critical inclusion of diverse groups in AI development and governance to foster more democratic and socially responsible technologies.

## 1. Introduction

The rapid advancement of generative artificial intelligence (AI) has raised pressing questions about equity, diversity, and inclusion (EDI) in algorithmic data curation and presentation. While generative AI offers significant potential for innovation across domains, I argue that it also inherits and amplifies human biases embedded in the data it is trained on (O'Neil, 2016; Benjamin, 2019). These biases pose a critical threat to democratic values when they result in the exclusion or misrepresentation of equity-deserving groups such as women, racial minorities, older individuals, and people with disabilities. As societies increasingly rely on AI-generated outputs in fields such as media, hiring, marketing, and education, it becomes ethically and politically urgent to understand how these technologies reflect, or distort, social diversity (Barocas and Selbst, 2016; Durrheim et al., 2023).

In this paper, I examine how generative AI systems visually represent individuals across a wide range of occupations, focusing on four key identity dimensions: gender, race, age, and visible disability. I draw on images produced by widely used generative AI tools and explore whether these technologies reinforce occupational stereotypes and structural inequalities, or they can serve as a platform for more inclusive and equitable representation. Prior research has shown that algorithmic systems, frequently replicate discriminatory patterns due to biased training datasets or opaque design processes (Lambrecht and Tucker, 2019; Chen, 2023; Bursell and Roumbanis, 2024). In particular, I consider how the visual and textual outputs of large language models (LLMs) and image generators may subtly reinforce dominant cultural norms while rendering marginalized groups invisible (Omiye et al., 2023; Salinas et al., 2023).

Framed within broader debates on algorithmic governance and democratic accountability (Helbing et al., 2019; Helberger, 2019), my study contributes empirical evidence on how generative AI tools represent, or fail to represent, human diversity in the labor market. The findings reveal consistent underrepresentation and stereotyping of specific social groups, raising concerns about the normalization of structural exclusion through AI-generated media. By focusing on the visual outputs of generative AI, a dimension often overlooked in fairness and bias research, I aim to expand our understanding of how these technologies shape social norms and perpetuate inequality. I conclude by recommending the implementation of algorithmic diversity exposure and fairness-aware design practices, and by urging policymakers, researchers, and technology developers to treat inclusion in AI as a core component of democratic governance.

## 2. Diversity in the workplace and democracy

Democracy, labor, and diversity have been frequently discussed in relation to one another, as they shape the socio-political and economic structures of modern societies (Sen, 2014). The treatment of workers within the workplace and the broader economy by governments and capitalists has long been a central focus in debates on political-economic schools such as capitalism and socialism (Harvey, 2005).

Moreover, diversity in the job market, particularly regarding the roles of women, people of color, and other marginalized groups, has been widely discussed as an essential manifestation of democracy (Young, 2002). A democratic society, in principle, promotes equal opportunities and fair treatment in employment; however, systemic barriers continue to hinder full participation by underrepresented groups (Acker, 2006). Studies have shown that workplace discrimination, wage gaps, and glass ceiling effects persist despite legislative efforts to ensure inclusivity and equity (Reskin, 2000).

In this vein, it is crucial to examine how biases, stereotypes, and structural inequalities in the job market influence the way AI perceives human labor. AI systems, trained on historical employment data, often inherit and perpetuate biases present in human decision-making, exacerbating existing inequalities (O'Neil, 2016). Research indicates that algorithmic hiring tools can disadvantage minority candidates due to biased training datasets, leading to exclusionary hiring practices (Barocas and Selbst, 2016). Furthermore, AI-driven workplace monitoring and automation have raised concerns about labor displacement and the potential deepening of socio-economic inequalities, particularly for workers in precarious employment (Eubanks, 2018). Therefore, studying the intersection of AI, labor bias, and workplace diversity is of critical importance for ensuring that technological advancements align with democratic values and promote equitable labor practices (Benjamin, 2019).

## 3. Mutual impacts: technology and democratic values

Technology is a human creation and, as such, inherently reflects our values, prejudices, and biases. Additionally, it plays a crucial role in shaping societal norms and social contracts. As Nemitz (2018) notes, we live in a world shaped by technology at least as much as by law. Just as people influence the law and the law guides human behavior, we need also to recognize—and actively engage in—the mutual shaping of law and technology.

The positive affordances of technology include, but are not limited to, connectedness and self-expression on social media, supporting open, transparent, and accountable governments, and facilitating more inclusive decision-making and democratic governance. Moreover, technology can have a positive impact on diverse groups by helping them gain skills, cultural competencies, and human and social capital. However, at the same time, digital technologies—particularly social platforms and generative AI—are increasingly seen as a critical threat to democracy. Big tech companies control AI, social media, and data, and democratic governments have limited control over algorithms. Political actors use algorithms and bots for manipulation and social control, while authoritarians and extremists use technologies for disconnection, censorship, misinformation, and surveillance to maintain power structures.

Policymakers and political scientists are increasingly discussing societies are transforming under the effects of the Internet, social media, AI, and algorithms. However, it remains unclear how democracies will evolve and how current governance structures will adapt (Helbing et al., 2019). Without a doubt, future democracies will require algorithms that uphold democratic principles.

Accordingly, four types of recommender algorithms—liberal, participatory, deliberative, and critical—with varying levels of inclusion and participation have been conceptualized (Helberger, 2019). The liberal algorithm focuses on individual autonomy, self-development, and the dispersion of power, primarily serving users' preferences by recommending content based on their past behaviors and interests. It assumes that citizens should be free to choose the information they consume, emphasizing personal relevance over exposure to diverse perspectives. In contrast, the participatory algorithm is rooted in participatory democracy and prioritizes inclusiveness, active citizenship, and civic engagement. It aims to present a broad range of perspectives and encourage participation by making users aware of different viewpoints rather than simply catering to their preferences. The deliberative algorithm, aligned with deliberative democracy, values open-mindedness, tolerance, and public discourse. Its goal is to foster informed debate by exposing users to diverse and conflicting viewpoints, actively challenging their existing beliefs to promote critical thinking and reduce ideological polarization. Finally, the critical algorithm takes a more activist stance, deliberately amplifying marginalized voices and challenging dominant narratives. It prioritizes content that disrupts mainstream perspectives, defies prejudices, and encourages critical reflection, aiming to create a more inclusive public discourse.

These four types of algorithms provide the technical possibility of including diverse groups. Yet, there are limited opportunities for individuals to influence these algorithms. Currently, people can only use or select from options provided by existing algorithms, while proposed frameworks primarily focus on embedding more diverse choices within algorithmic nudges—designed to subtly nudge users and influence their choices, behaviors, or decision-making processes by shaping how information is presented or recommended—rather than empowering individuals to develop and define their own algorithms.

If we assume that people are given the opportunity to play a role in inscribing algorithms, the next critical question is whether the conflicting interests of people result in democratic algorithms or perpetuate current human biases, stereotypes, divides, and challenges such as extremism, racism, and polarization. If algorithms reflect societal biases, they could be seen as democratic from one perspective, as democracy represents majority views rather than guaranteeing impartiality. However, critical democracy and critical algorithm studies argue that minority groups should be equally represented, regardless of their population size. For example, when a Black woman asks a generative AI to depict a university professor, she should have the same chance of seeing a Black woman as a white man. The recent rise of generative AI and LLMs is a good opportunity to explore this question, since the LLMs are being developed based on human language and the underlying data is primarily human generated data that could potentially show how the crowd's opinions affect AI and consequently our democratic values. LLMs risk propagating racism, as these models are built on natural language data that reflect our values, biases, and prejudices (Salinas et al., 2023). These language models perpetuate race-based medicine (Omiye et al., 2023), express gendered stereotypes about occupations (Kotek, Dockum and Sun, 2023), and may even equate terms like "immigrant" and "refugee" as synonyms, leading to associations with terms like "illegal" in different paraphrased outputs (Durrheim et al., 2023). These intersectional biases can have unpredictable impacts on our democracies. Therefore, policymakers should urgently audit and regulate AI based on its potential risks. In this essay, I underscore this need by quantifying some of the non-democratic biases present in AI tools.

## 4. How does AI see humans in their occupations?

Equity, diversity, and inclusion in the workplace and hierarchy of decision making is one of the main manifestations of democratic societies. Algorithms influence how people organize socially and how opportunities and power are distributed in workplace and the wider society. For instance, gender-based algorithmic bias in the display of science, technology, engineering and mathematics (STEM) job advertisements is apparent (Lambrecht and Tucker, 2019). This bias is not limited to gender; algorithmic bias based on race, color, and personality traits is also documented (Chen, 2023). Even when organizations aim to leverage unbiased algorithms to reduce discrimination, diversity may decrease due to unintended and emergent human-machine interactions (Bursell and Roumbanis, 2024). In this context, bias can be imposed by algorithms. Generative AI adds another risk factor: datasets biased by human generated data, as mentioned earlier. To uncover this risk, I experimented with three generative AI tools and generated 444 images across 37 occupations. Here, I briefly present the results.

## 5. Data and results

The underlying logic of AI image generators involves inferring meaning from text based on LLMs and then converting that text into visual outputs. Before launching the main experiment, I conducted a preliminary experiment with ChatGPT and Meta AI, both of which provide textual and visual outputs, to assess whether any noticeable bias exists in textual outputs compared to visual ones. I used a straightforward prompt to generate images of various professions, and the results were surprising. The prompt I used was: "How should a taxi driver's appearance be? Please give me a picture."

ChatGPT generated an image after generating the text below:

*"A taxi driver's appearance should be professional and approachable to create a good impression on passengers. Here's a description:*

- *Clothing: A neat and clean shirt, preferably with a collar. The shirt could be a solid color like white, light blue, or light gray. The driver might also wear a jacket, especially if the weather is cooler. Dark-colored pants or jeans, free of wrinkles and stains, are ideal.*
- *Footwear: Comfortable but presentable shoes, such as loafers or dress shoes, are preferred. Sneakers might be acceptable if they are clean and in good condition.*

- *Grooming: Well-groomed hair, trimmed facial hair (if any), and general cleanliness are important. The driver should avoid strong fragrances or excessive cologne.*
  *Let me generate an image based on this description."*

The text appears to be gender-neutral, age-neutral, and race-neutral. However, the first generated image depicted a young white man. When I requested another image, it again generated a young white man (Figure 1).

**Figure 1. ChatGPT's DALL.E generated two young white men in response to two consecutive prompts to show how a taxi driver's appearance should be**

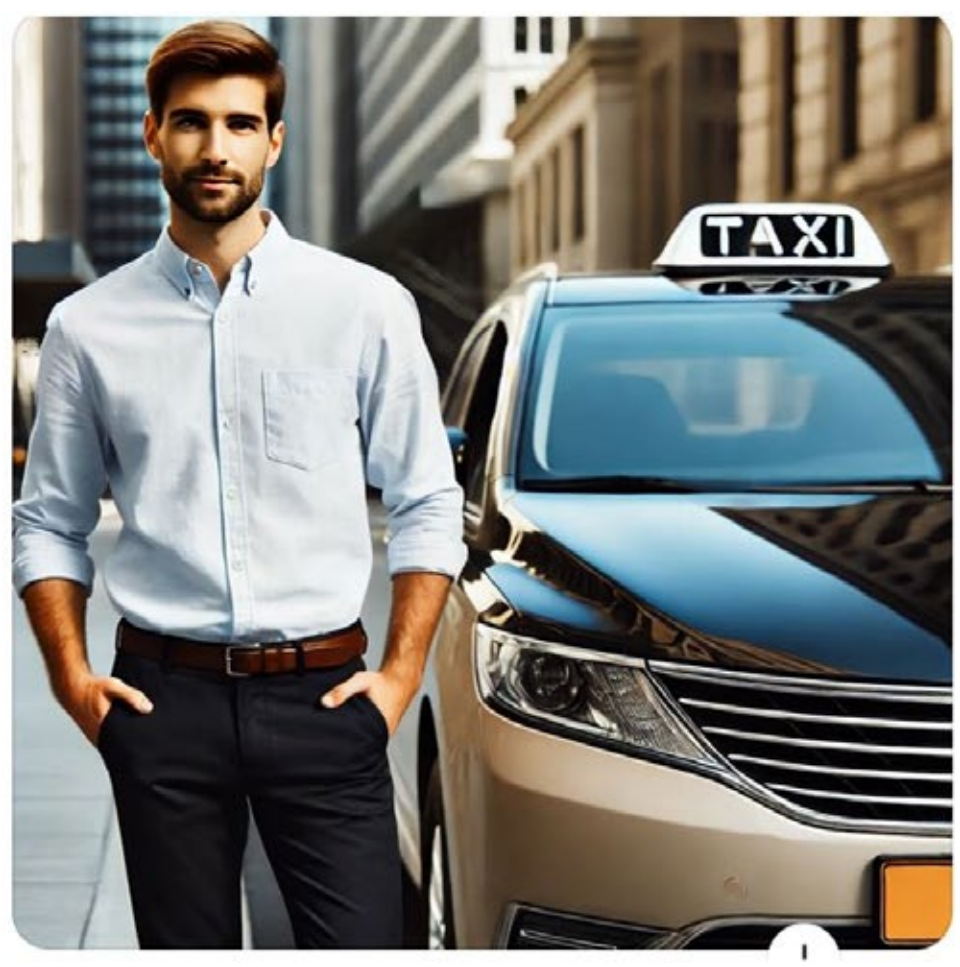

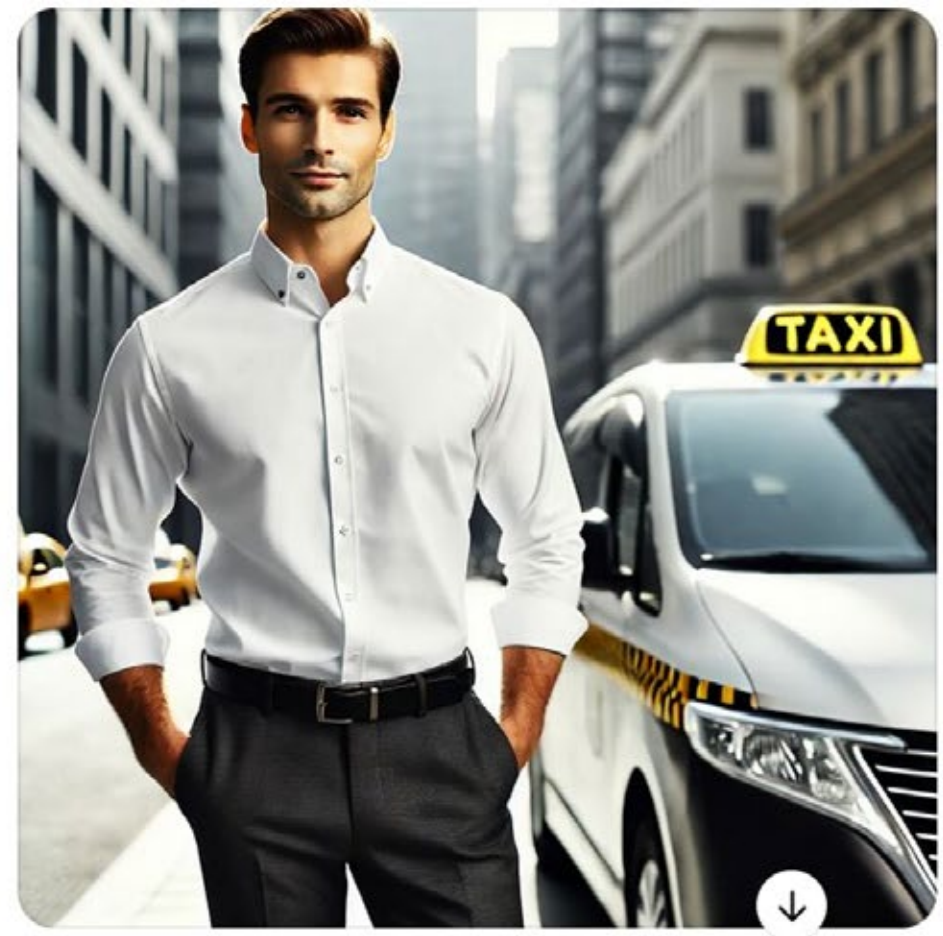

In response to the same prompt, Meta AI generated four images at once (Figure 2). However, it did not generate any text. I asked it to generate a text in response to this prompt: "Would you please describe in text how the appearance of a taxi driver should be?"

Meta AI responded:

*"A taxi driver's appearance should be:*
*Professional:*
- *Clean and ironed uniform or attire*
- *Company logo or identification visible*
- *Name tag or ID badge*

*Well-groomed:*
- *Clean-shaven or neatly trimmed beard*
- *Hair tidy and away from face*
- *Nails clean and trimmed*

*Comfortable and practical:*
- *Comfortable shoes for long drives*
- *Minimal jewelry or accessories*

- *No strong perfumes or colognes*

*Approachable and friendly:*

- *Smile and make eye contact*
- *Neat and clean vehicle interior*
- *Helpful and courteous demeanor*

*Remember, appearance can vary depending on local regulations, company policies, and personal style. However, a professional, well-groomed, and approachable appearance is generally expected of taxi drivers."*

**Figure 2. Meta AI: how a taxi driver should appear**

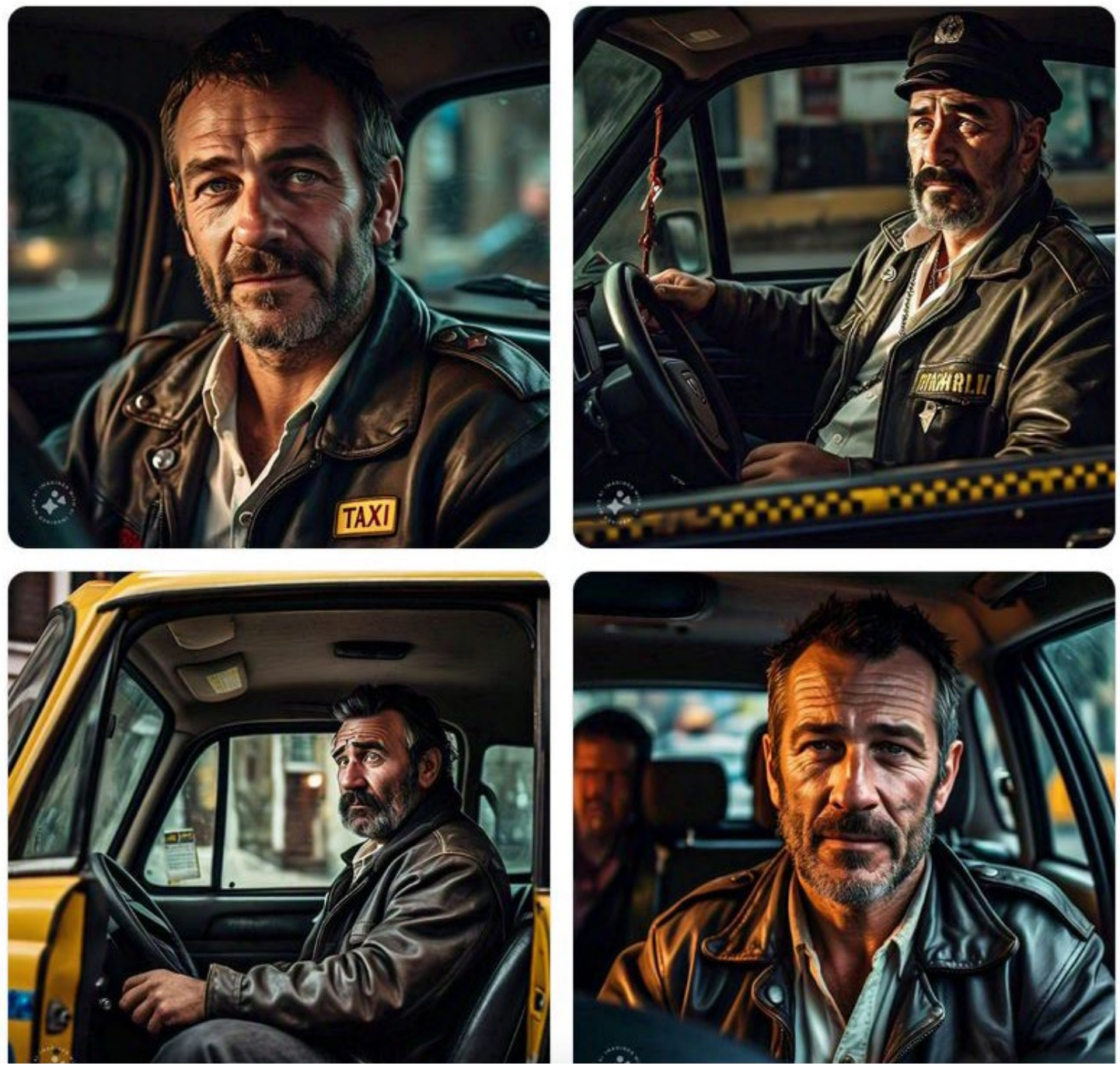

Both the textual and visual outputs of Meta AI appeared biased and skewed toward masculinity and I saw some mismatch between the text and the images. These preliminary findings prompted me to conduct a broader experiment across a larger sample of occupations. I examined three generative AI applications—Microsoft Designer, Meta AI, and Ideogram—by generating 444 images based on a purposefully selected sample of 37 occupations. Before presenting the results, I must highlight two important points. As the author of this essay, I felt excluded from all occupations, as none of the images reflected my race. Additionally, I did not observe anyone with a visible disability.

### *5.1. Stereotypes*

The representation of women and middle-aged and elderly individuals in AI-generated images of various occupations seems to reinforce existing stereotypes or mirror their current status in the job market (Figure 3). From a critical democracy perspective, even the latter root cause is not also truly democratic, as the job market itself is not necessarily democratic. While a minority group making up 1% of the population may also represent 1% of the workforce, generative AI outputs should not reflect this proportion. Instead, they should ensure equal representation across diverse groups in society, rather than merely mirroring existing inequalities or realities.

**Figure 3. Age and gender stereotypes: sample occupations**

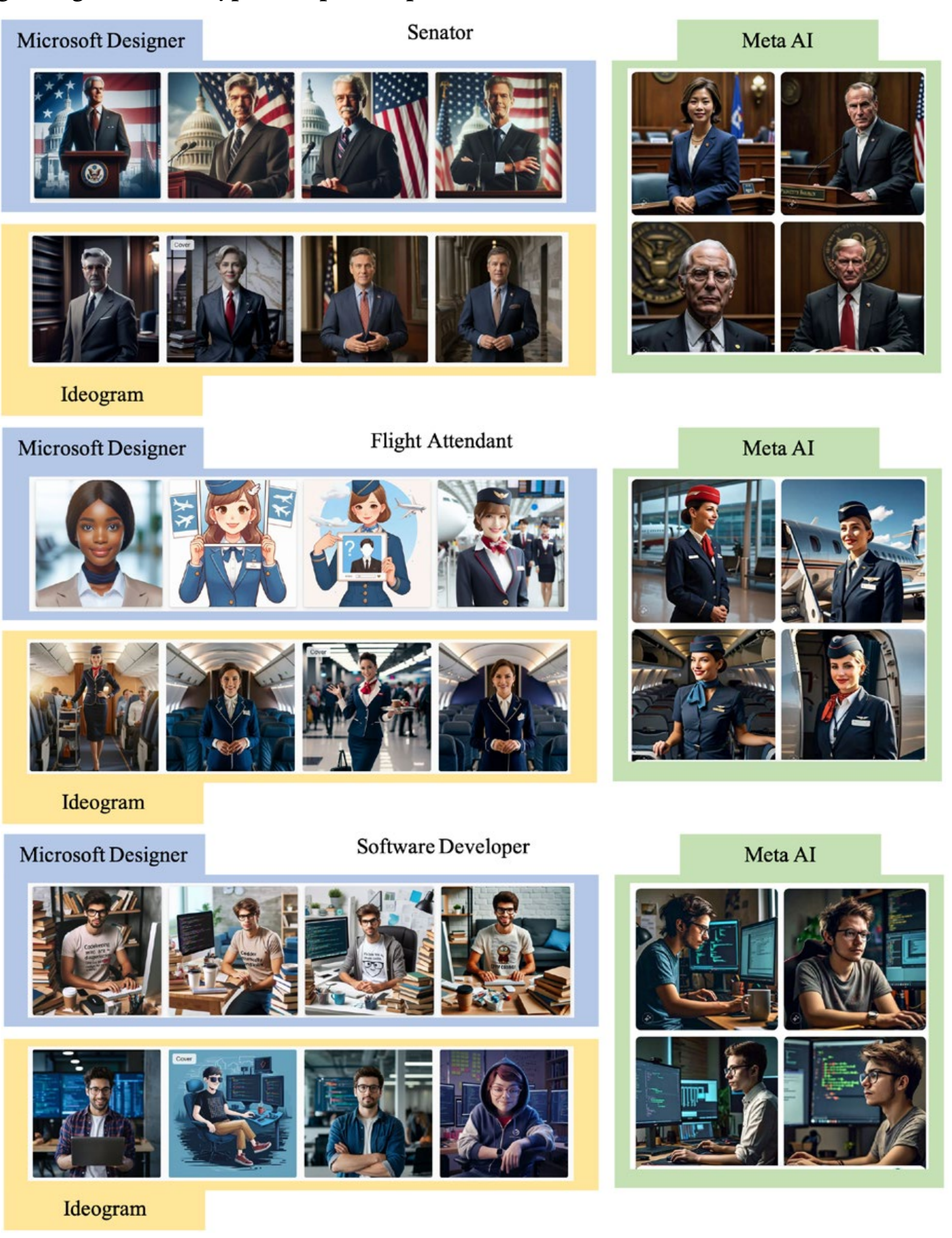

### *5.2. Women's representation*

Women are underrepresented in senior roles and technology jobs, equitably represented in advisory roles, and slightly overrepresented in the health and biotech sectors. I compared 10 managerial and senior roles—CEO, CTO, entrepreneur, mayor, senator, chef, neurosurgeon, physician, professor, and flight pilot— with 10 operational and junior roles—cashier, cleaning worker, farmer, flight attendant, laborer, nurse, receptionist, retail salesperson, secretary, and taxi driver—and found women account for 16% of managerial and senior roles while they make up 57% of junior and operational roles. I also observed how women are represented in different sectors. They account for 12% of technology occupations, 56% of healthcare and biotechnology jobs, and 50% of advisory roles including career advisor, financial advisor, life coach, lawyer, and realtor. Overall, women account for 42% of all generated images across all 37 occupations.

### *5.3. Black individuals' representation*

According to the data, Black individuals are significantly underrepresented across all occupations and sectors in the generated images, accounting for only 8% overall. They make up 5% of managerial and senior roles and 8% of junior and operational positions. In technology jobs, they represent 10%, in healthcare and biotechnology roles 12%, and in advisory positions 8%.

### *5.4. Middle-aged and elders' representation*

As previously discussed, age stereotypes are evident in the generated images. Young people account for 68% of all generated images. Middle-aged and elders account for 52% of senior and managerial roles and 25% of junior and operational roles. In technology roles, they represent 6%, in healthcare and biotechnology 26%, and in advisory positions 43%.

When comparing AI tools, Ideogram demonstrates slightly less bias than Microsoft Designer and Meta AI. Overall, the representation of equity-deserving groups varies across sectors and AI tools (see Table 1). To gain a clearer understanding of diversity in generative AI, I recommend analyzing the results occupation by occupation (see in the supplementary data). Aggregating different occupations in this context is neither reasonable for statistical analysis nor persuasive for mitigating non-democratic risks, as each occupation and group requires specific attention.

**Table 1. Representation of diverse groups in AI-generated images across occupations**

| Sector/tool | Women | Black individuals | Middle-aged and elders | Visible disability |
|---|---|---|---|---|
| Senior and Managerial (10 occupations) | 16% | 5% | 52% | 0% |
| Junior and managerial (10 occupations) | 57% | 8% | 25% | 0% |
| Technology excluding biotech (4 occupations) | 12% | 10% | 6% | 0% |
| Healthcare and biotech (6 occupations) | 56% | 12% | 26% | 0% |
| Advisory (5 occupations) | 50% | 8% | 43% | 0% |
| Microsoft Designer (37 occupations) | 29% | 4% | 21% | 0% |
| Ideogram (37 occupations) | 61% | 16% | 31% | 0% |
| Meta AI (37 occupations) | 37% | 4% | 45% | 0% |
| All 37 occupations/all tools | 42% | 8% | 32% | 0% |

*Source*: author's own elaboration.

## 6. Conclusions and implications

The findings reveal that generative AI is not equitably inclusive regarding gender, race, age, and visible disability. While there are differences across AI tools in some occupations or sectors, several key patterns emerged: (1) Black individuals are significantly underrepresented across all occupations and sectors. (2) Women are underrepresented in senior roles and technology jobs, equitably represented in advisory roles, and slightly overrepresented in the health and biotech sectors. (3) The representation of women, middle-aged, and elderly individuals in various occupations appears to align with existing stereotypes or reflects their current status in the job market. (4) No one with visible disability is represented in any occupations. (5) Ideogram exhibits slightly lesser bias compared to Microsoft Designer and Meta AI.

I argue that, whether these biases stem from the perpetuation of human stereotypes within the underlying datasets and algorithms of generative AI, or reflect the realities of the job market, they risk reinforcing undemocratic biases and stereotypes. I suggest implementing algorithmic diversity exposure in generative AI to mitigate these risks.

The findings have several implications for corporate social responsibility (CSR) and ethical business practices. AI development companies should critically assess how homophily among their programming teams—often leading to a uniformity in perspectives and lived experiences— may contribute to the perpetuation of biases embedded in the technology. Moreover, they should critically test their underlying datasets against equity, diversity, and inclusion criteria and techniques such as fairness-aware machine learning and differential impact analysis (Binns, 2018; Mehrabi et al., 2021) to examine how processes such as the formation of ethnic enclaves, social alienation, gender socialization, racism, ageism, and tokenism are exacerbated by the stereotypes reflected in the AI-generated images observed in this research. By actively promoting an inclusive workforce and encouraging diverse contributions to the coding and training processes, AI companies can mitigate the unintentional reproduction of societal stereotypes and biases. Such initiatives will not only enhance the ethical standing of AI developers but also foster more inclusive AI technologies that better serve a diverse population.

For corporations integrating AI tools, testing for biases in generative AI and LLMs becomes crucial in maintaining an equitable and inclusive organizational culture. By evaluating how these AI systems may impact both their internal environment and their external customer interactions, companies can take proactive steps to identify and address potential risks. Biases in AI can not only reinforce existing inequalities within a workforce but may also alienate or disadvantage certain customer groups, impacting brand perception and customer loyalty. Therefore, companies adopting AI should incorporate routine audits and sensitivity analyses as part of their CSR strategies, ensuring that AI tools align with the organization's values of inclusivity, fairness, and social responsibility.

Supplementary Data (August 2024)

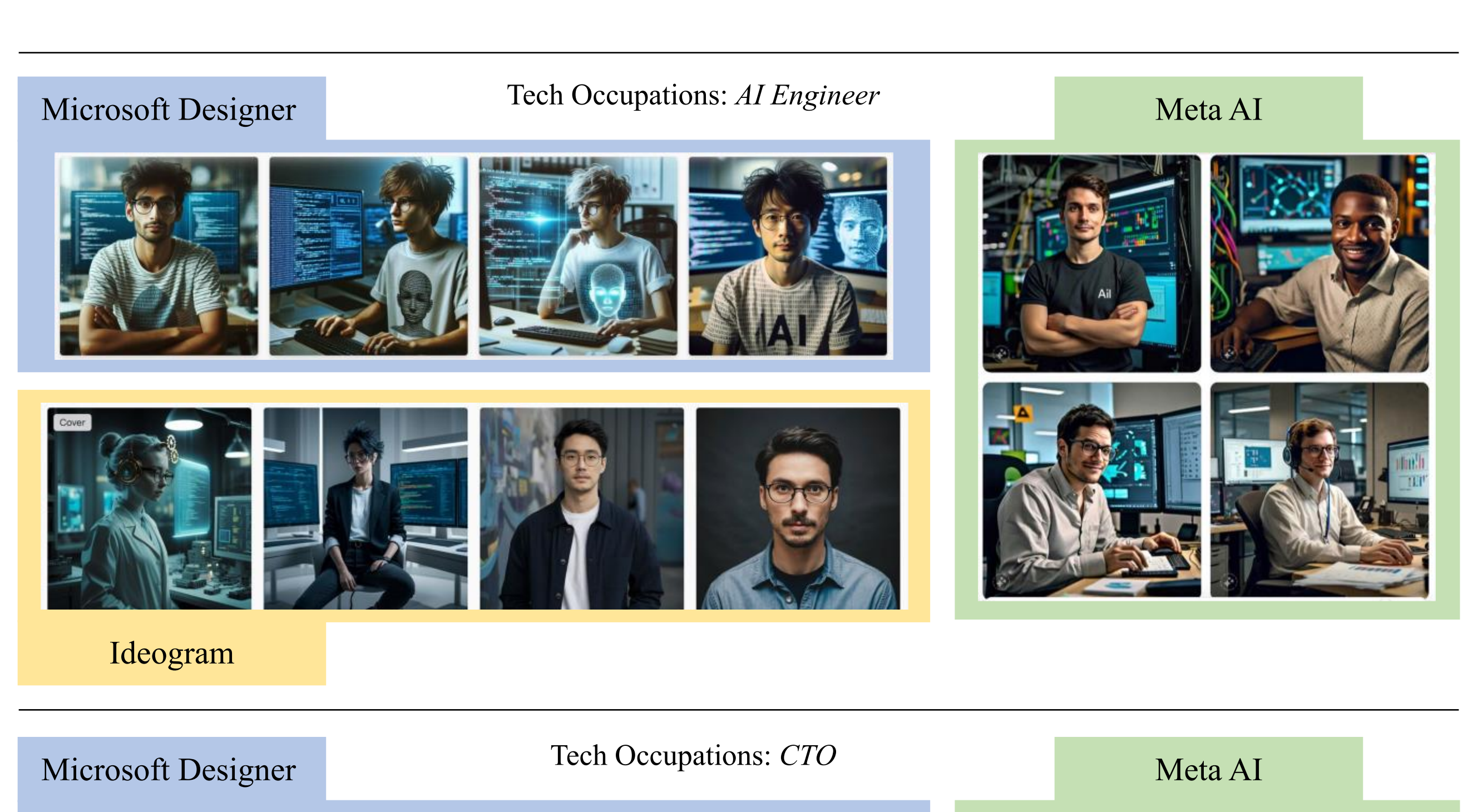

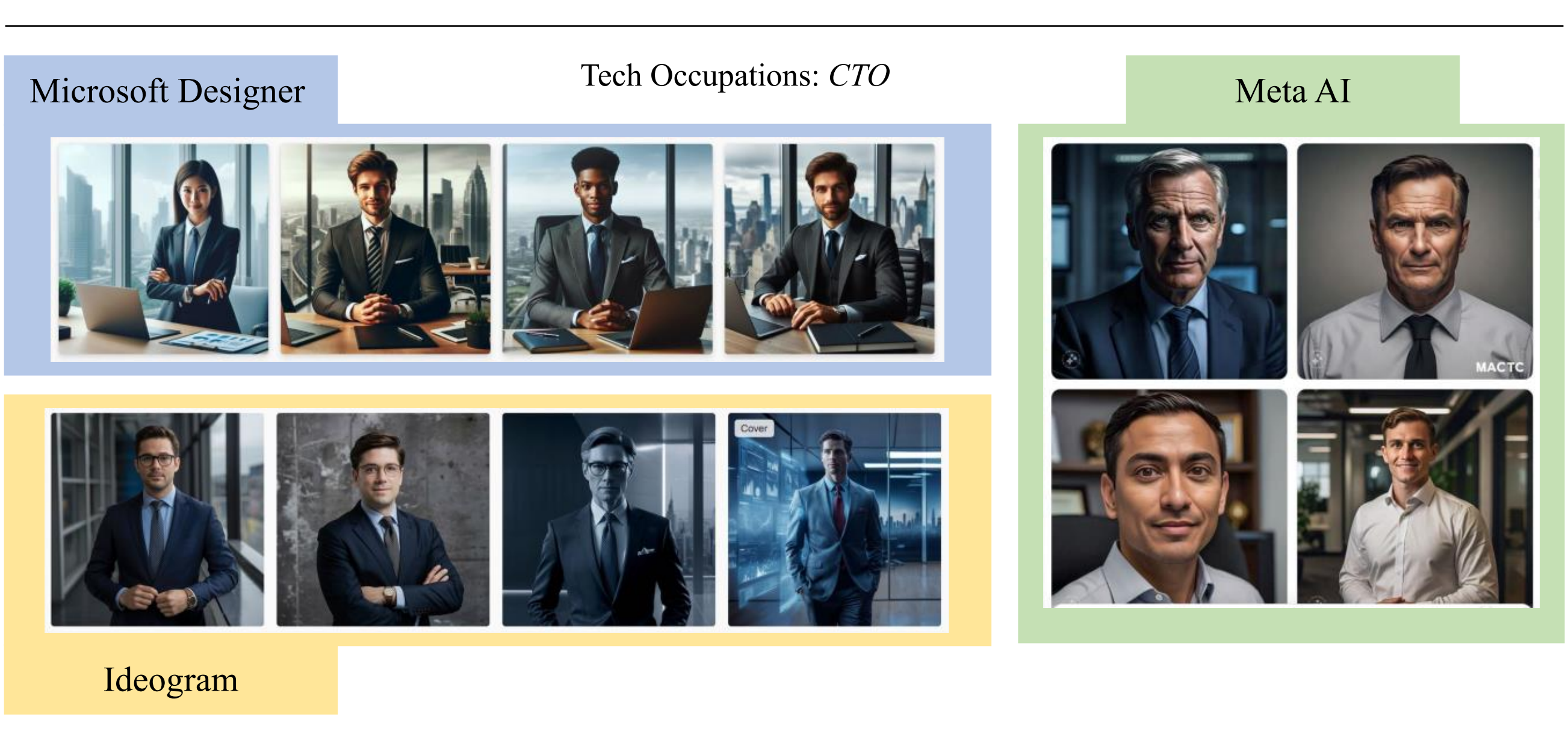

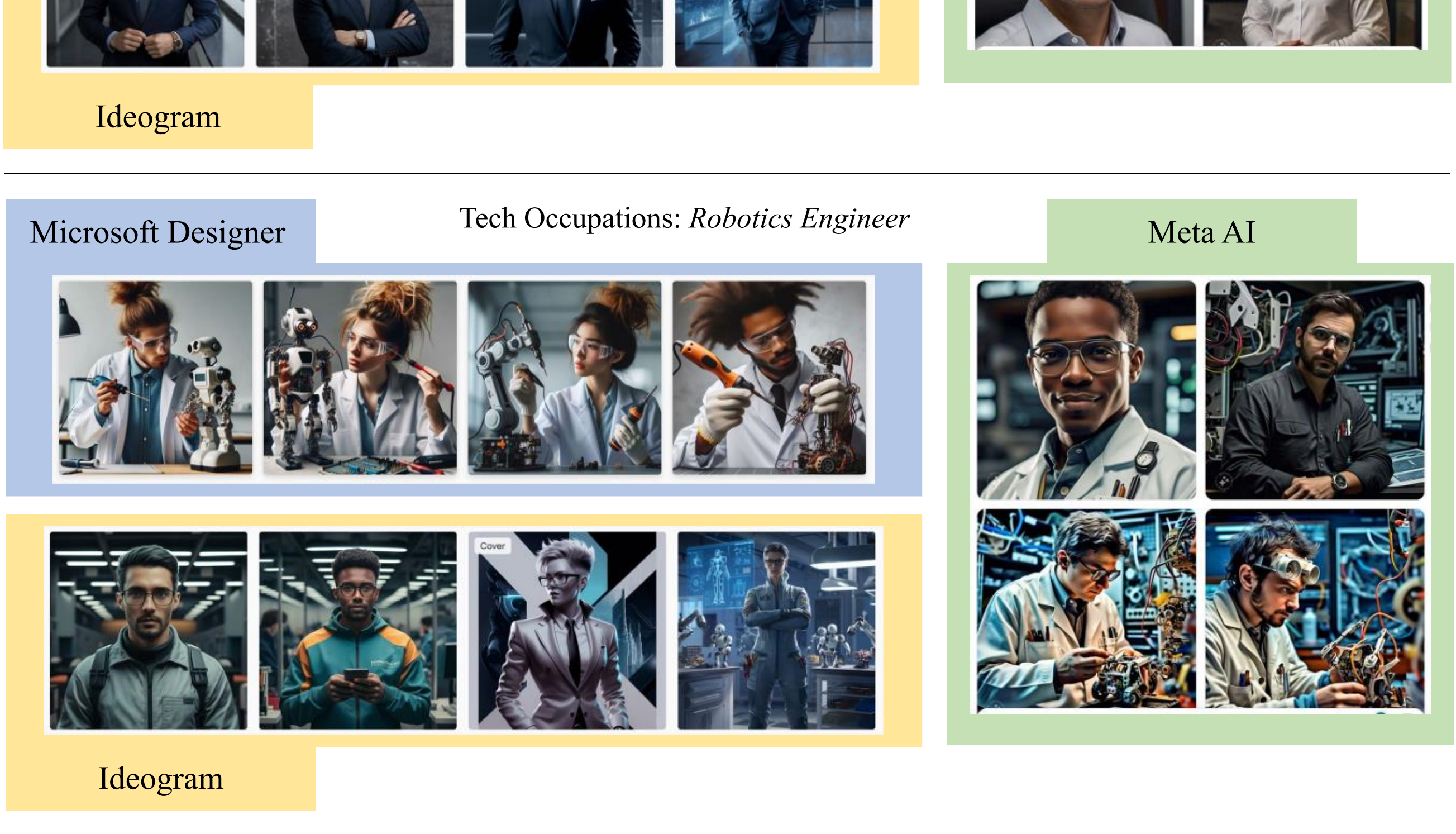

Microsoft Designer

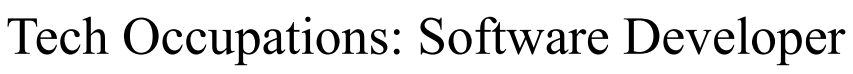

Meta AI

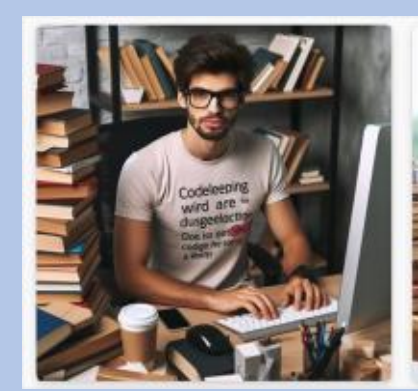
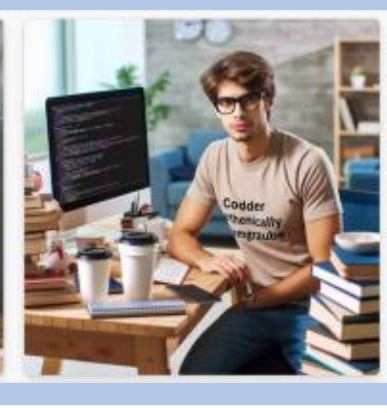
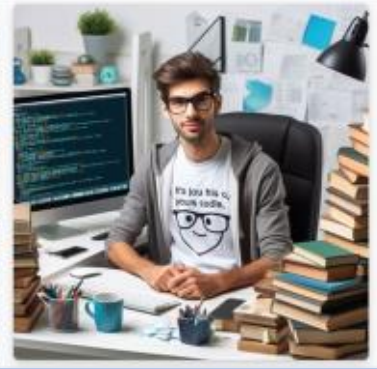
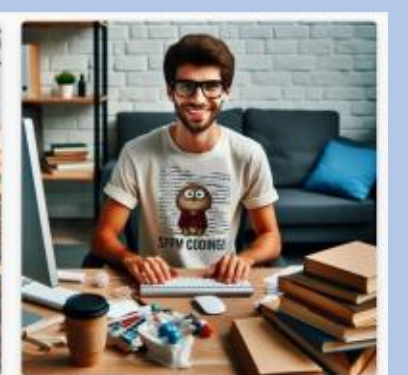
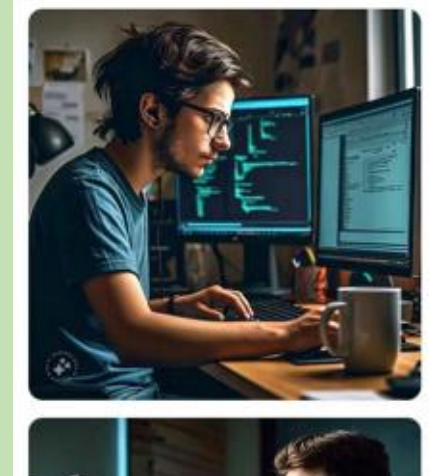
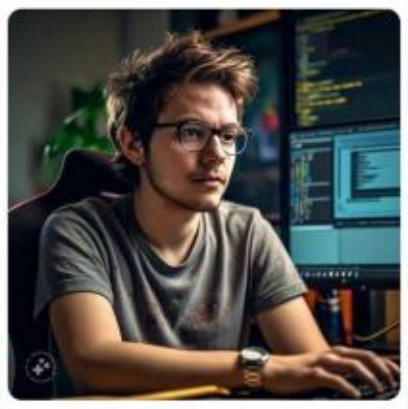
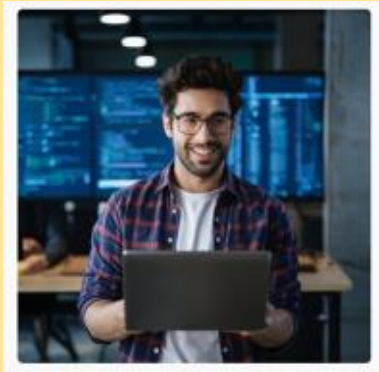
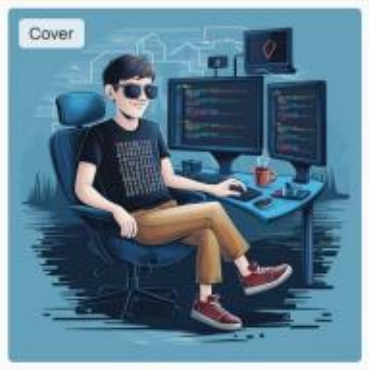
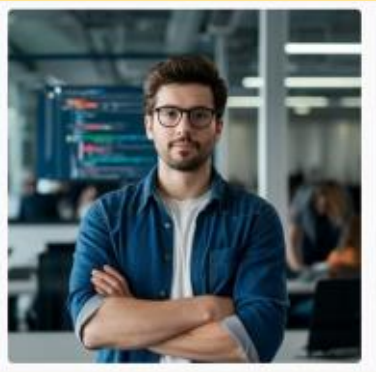
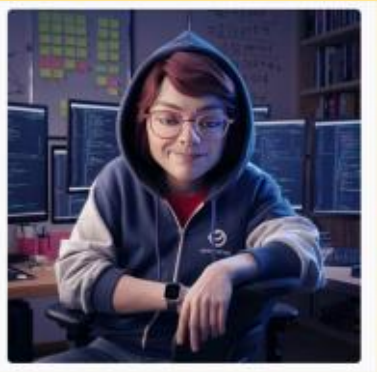
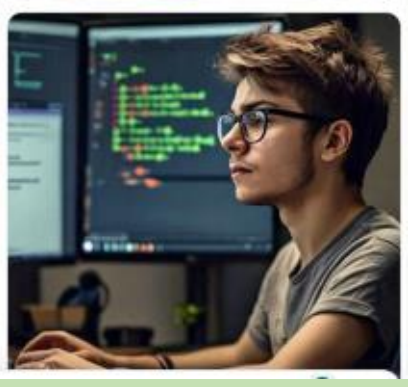

Ideogram

---

Microsoft Designer

Health/Biotech Occupations: *Biomedical Engineer*

Meta AI

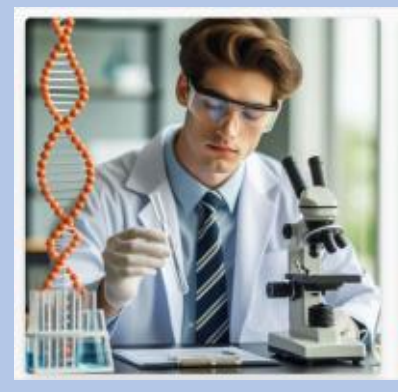
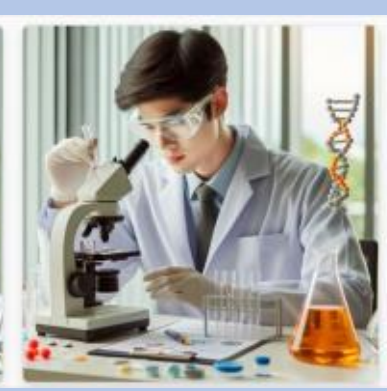
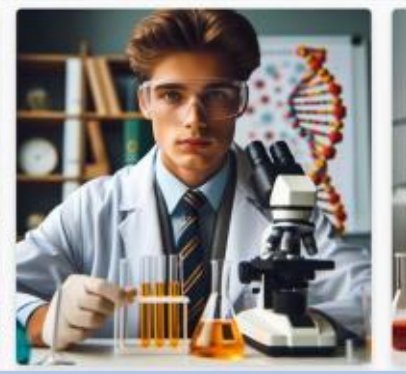
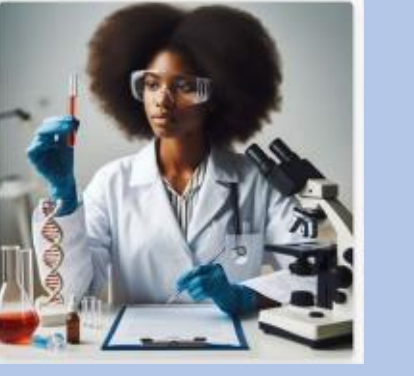
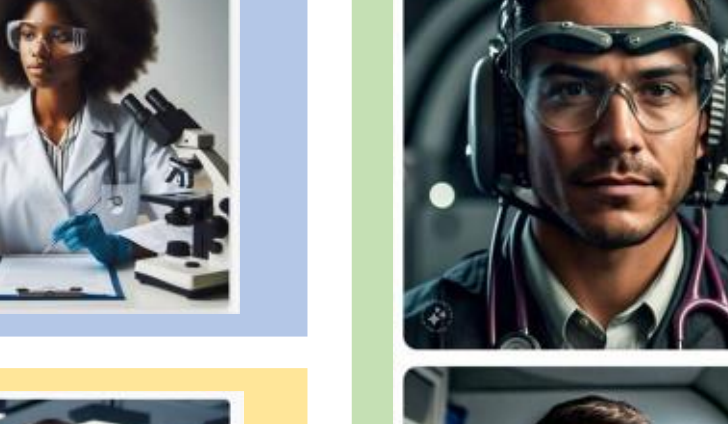
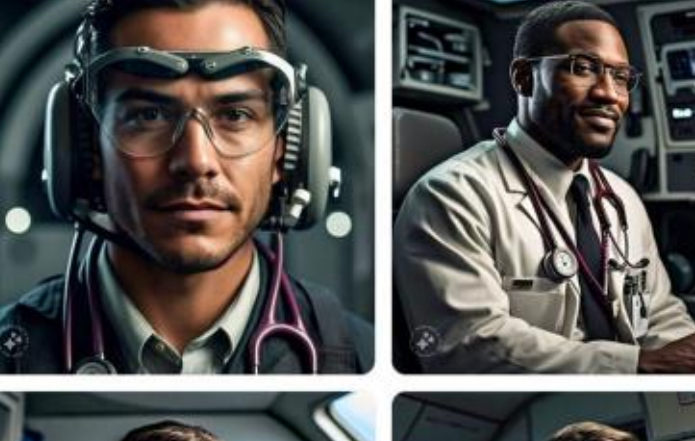
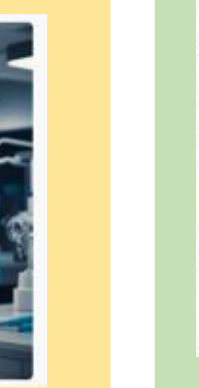
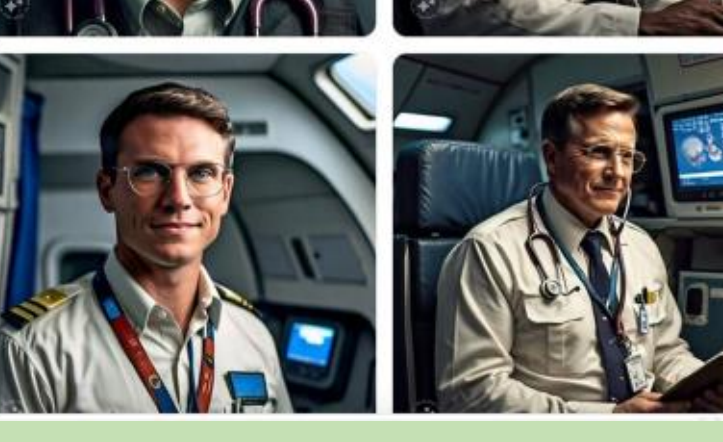

Ideogram

---

Microsoft Designer

Health/Biotech Occupations: *Biotechnologist*

Meta AI

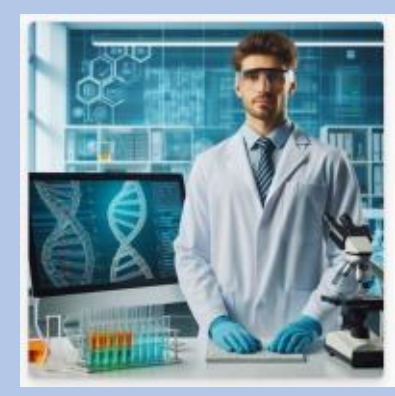
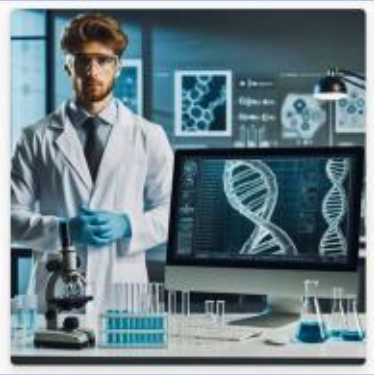
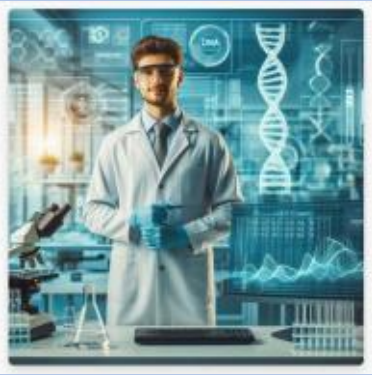
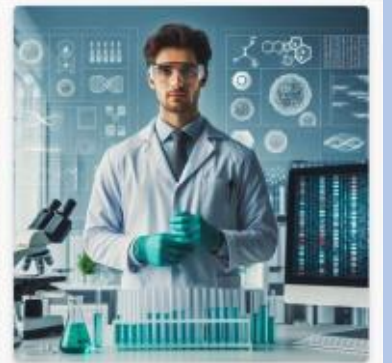
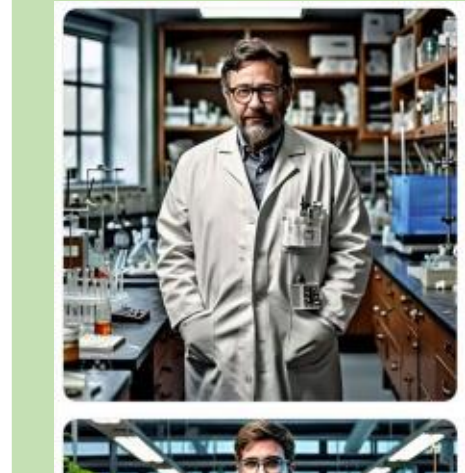
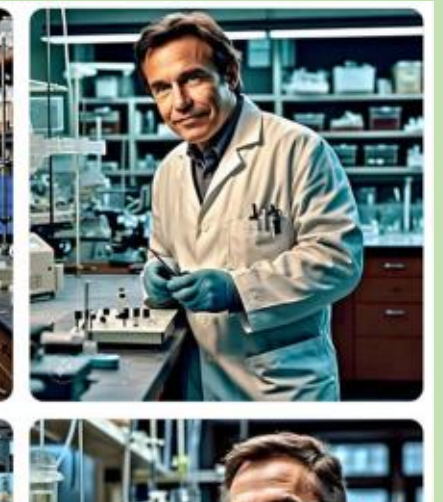
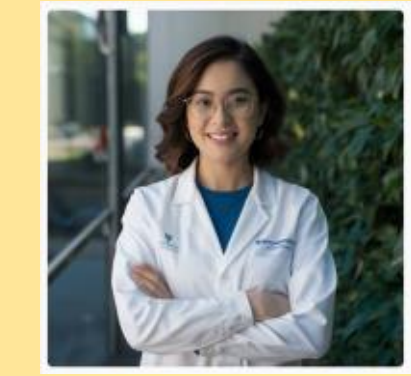
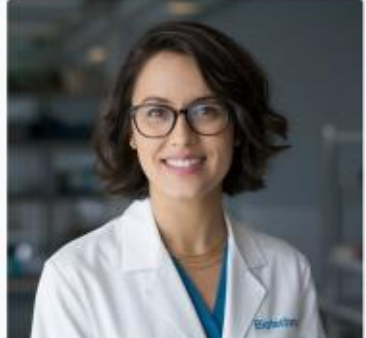
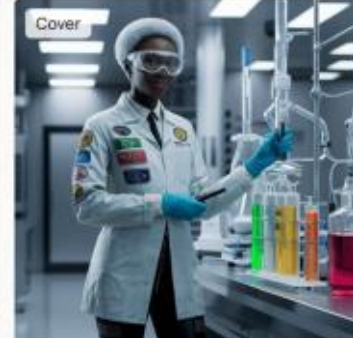
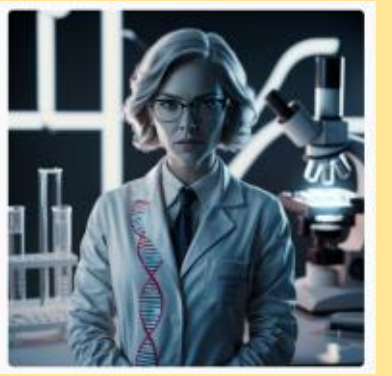
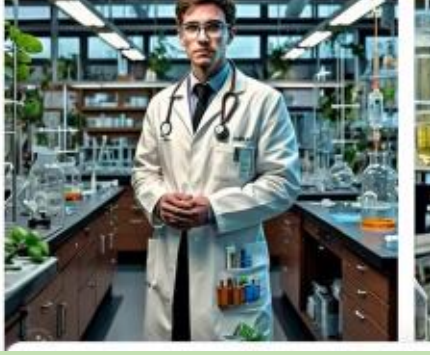
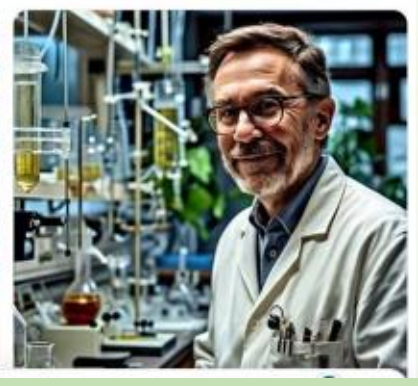

Ideogram

Microsoft Designer

Health/Biotech Occupations: *Healthcare Aide*

Meta AI

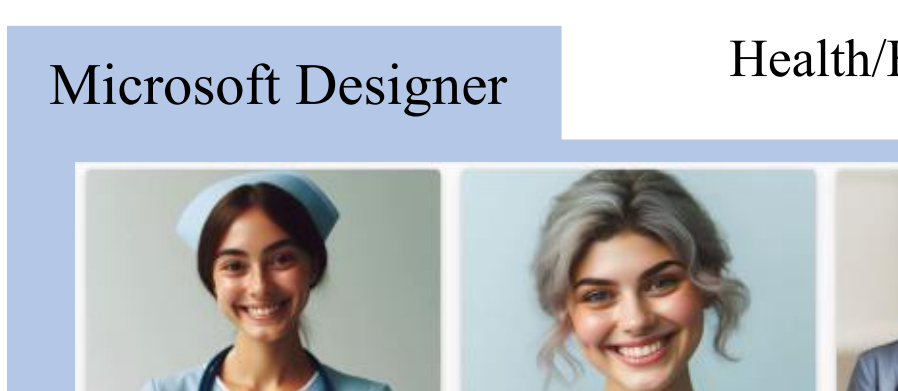

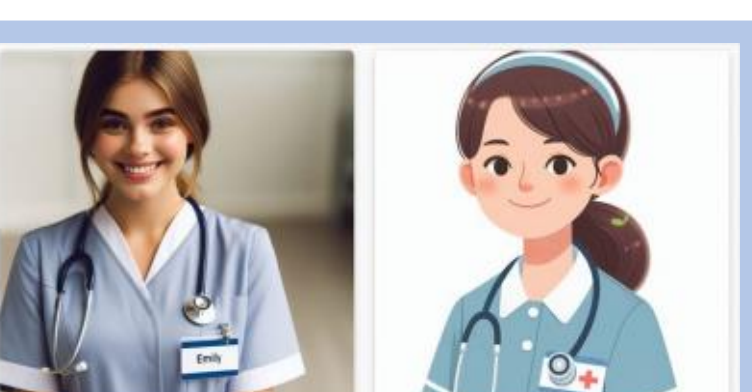

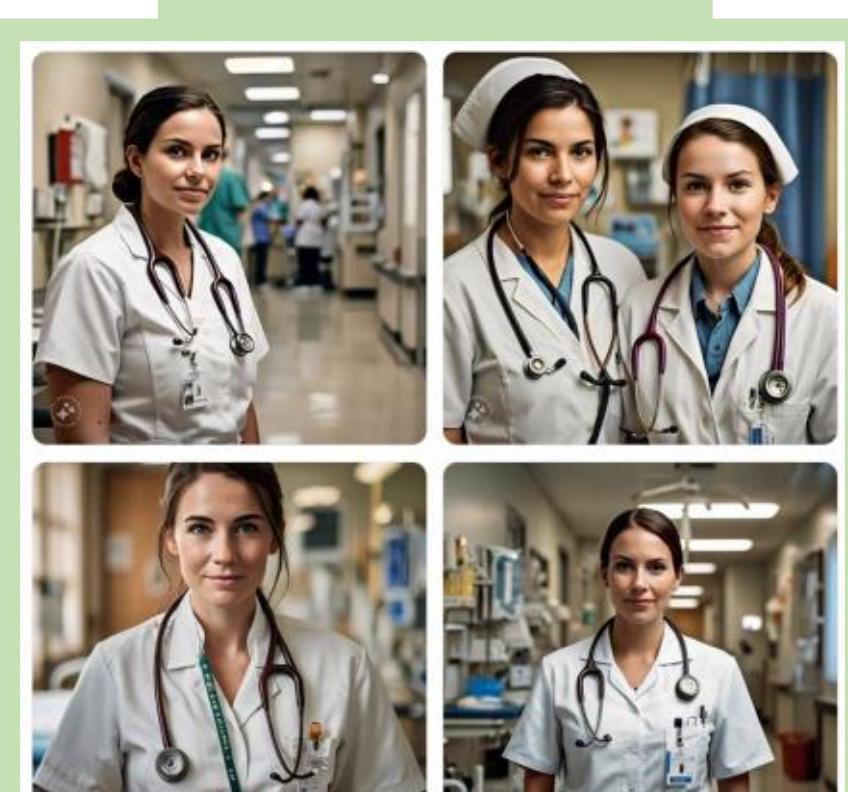

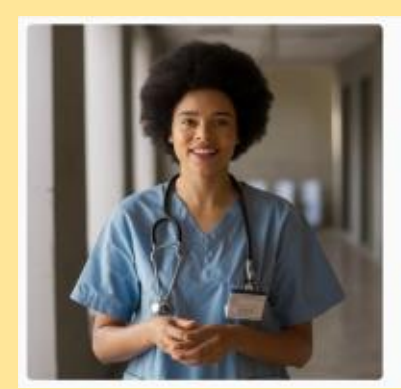

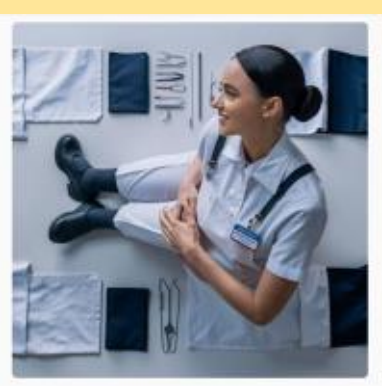

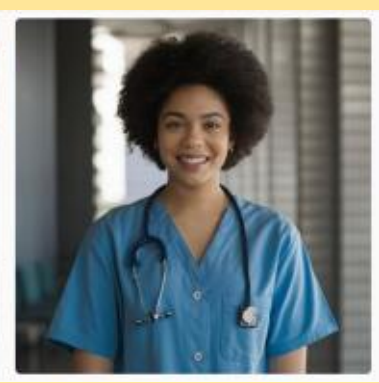

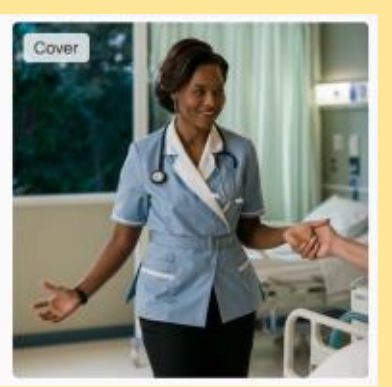

Ideogram

---

Microsoft Designer

Health/Biotech Occupations: *Neurosurgeon*

Meta AI

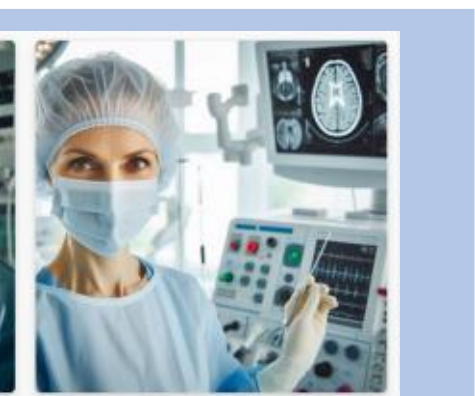

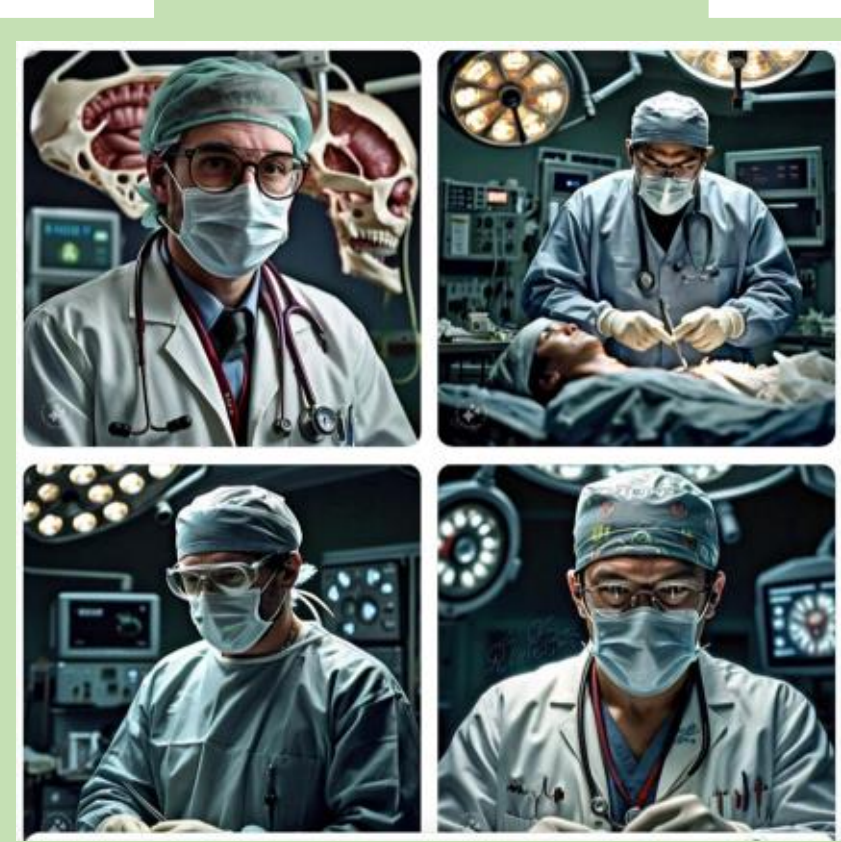

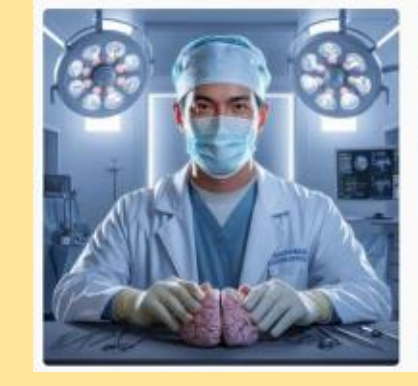

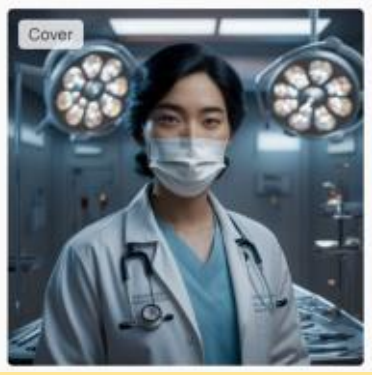

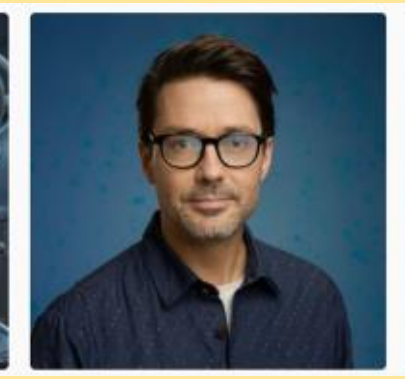

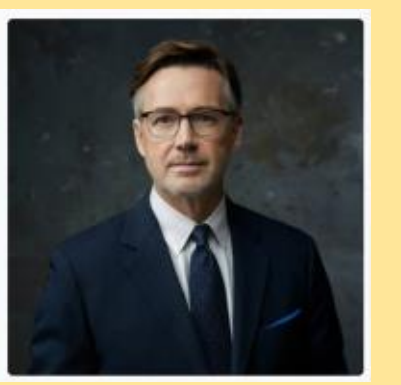

Ideogram

---

Microsoft Designer

Health/Biotech Occupations: *Nurse*

Meta AI

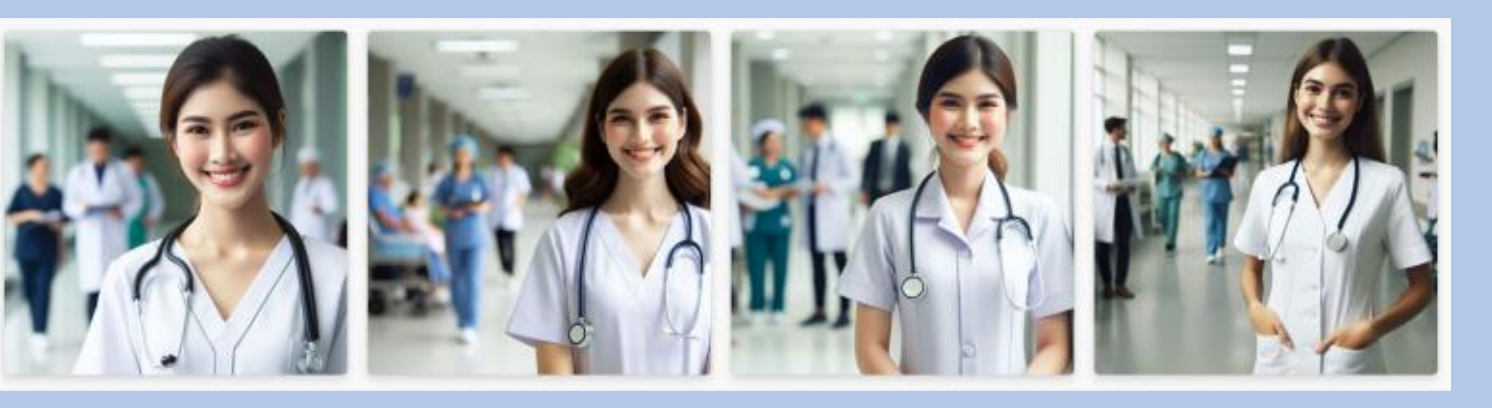

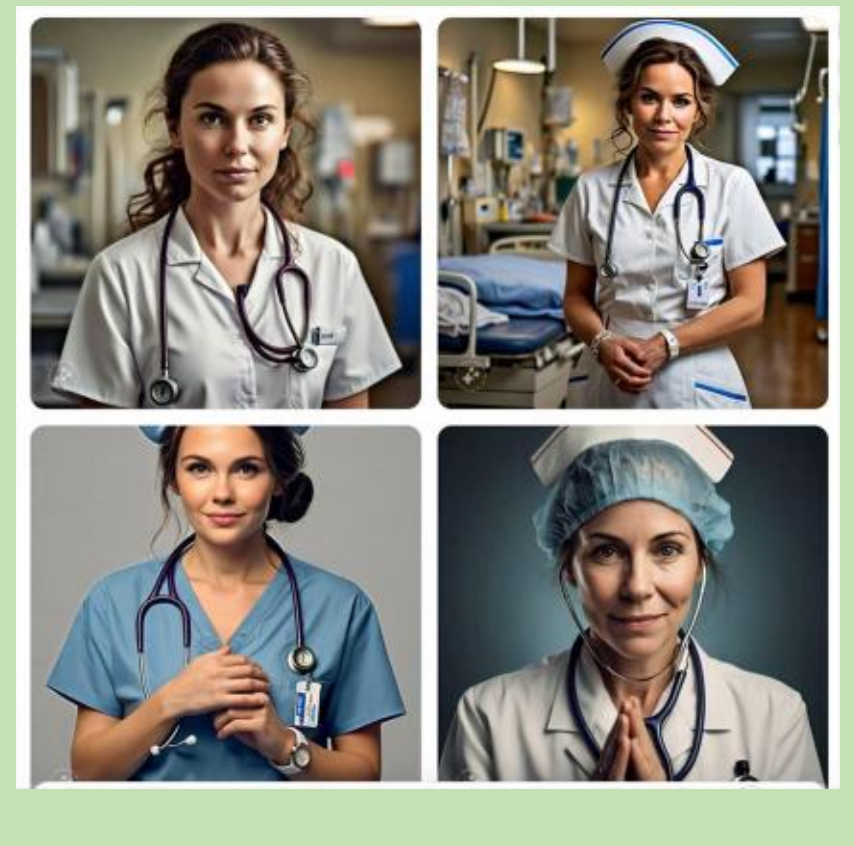

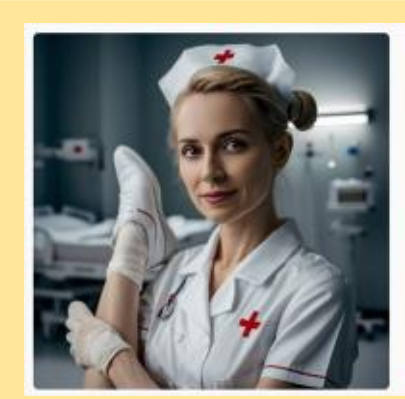

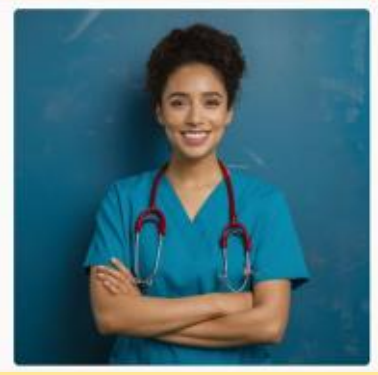

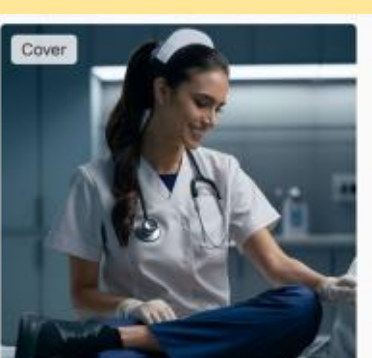

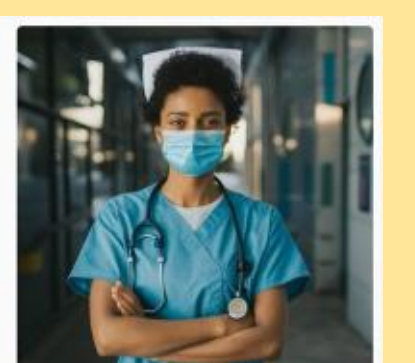

Ideogram

Health/Biotech Occupations: *Physician*

Microsoft Designer

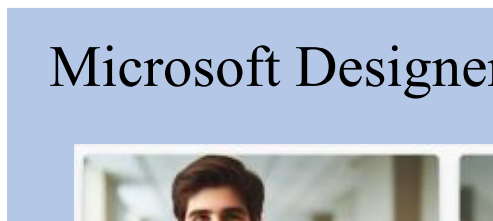
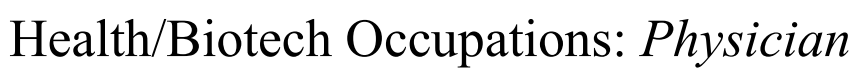
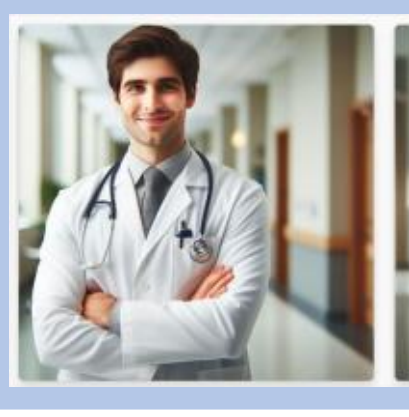
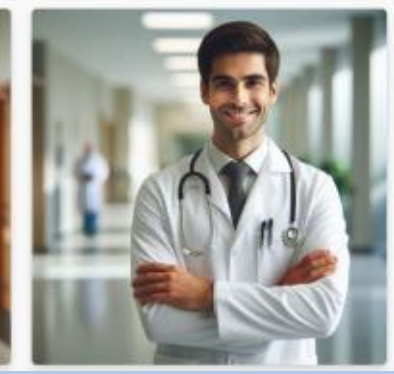
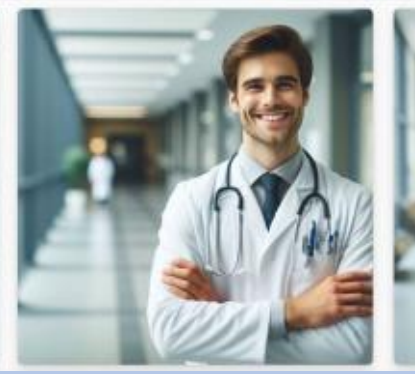
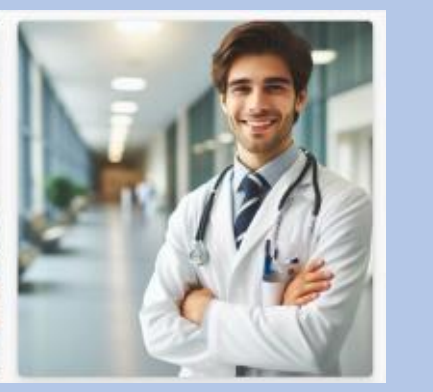

Meta AI

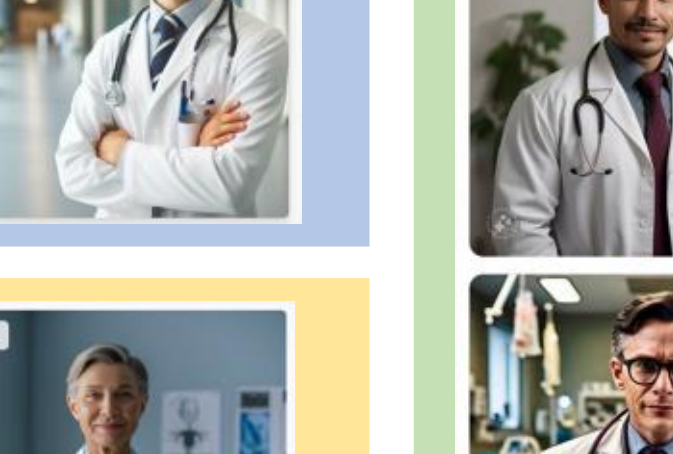
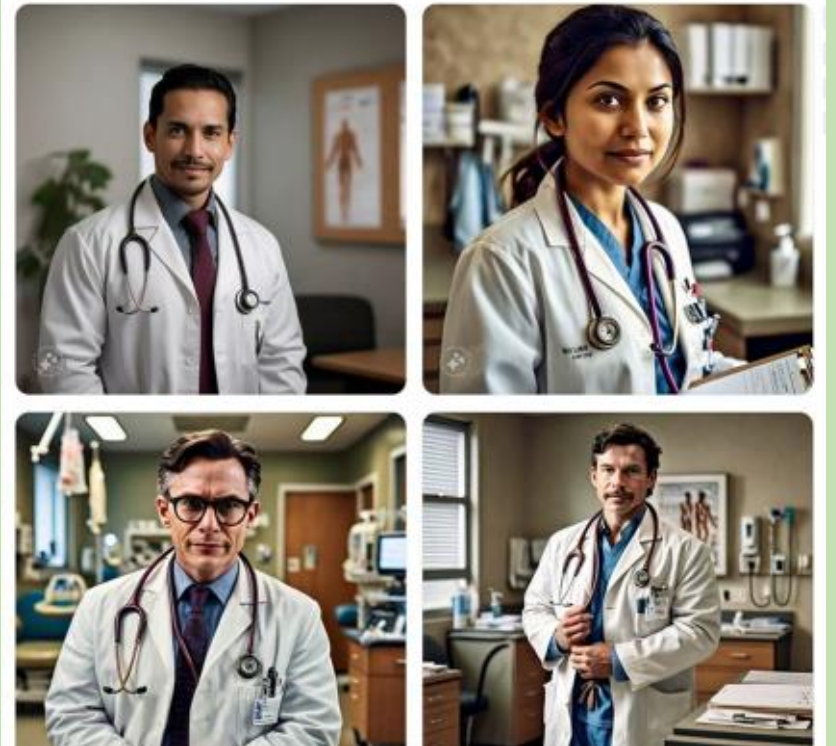

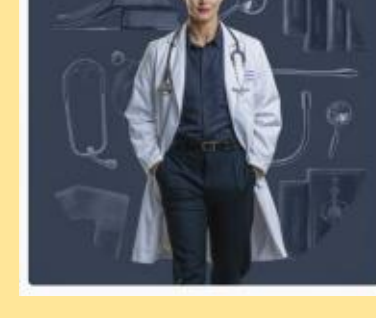
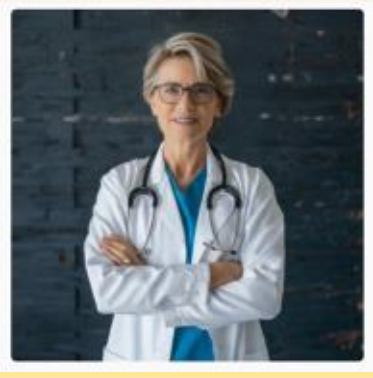
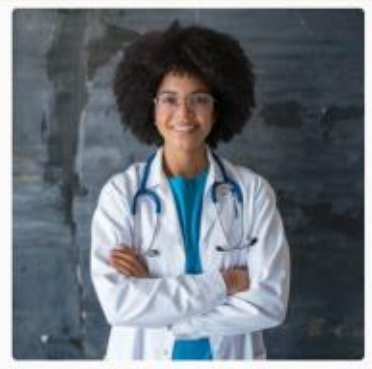
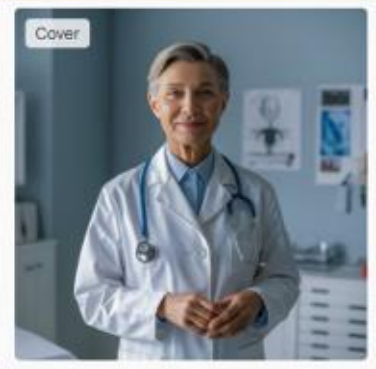

Ideogram

---

Advisory Occupations: *Career Advisor*

Microsoft Designer

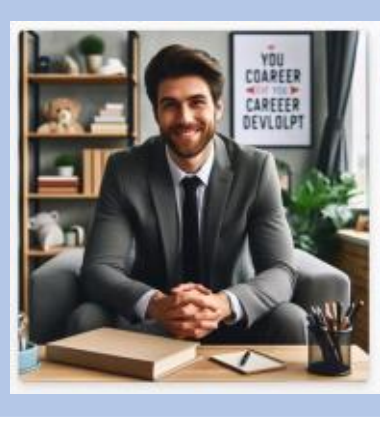
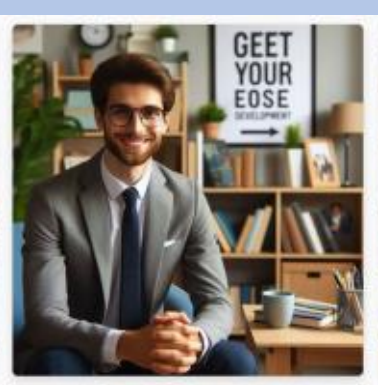

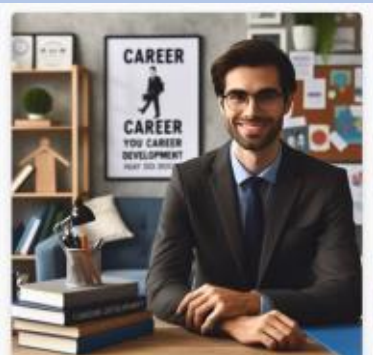

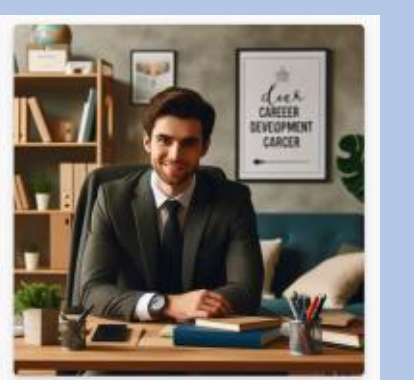
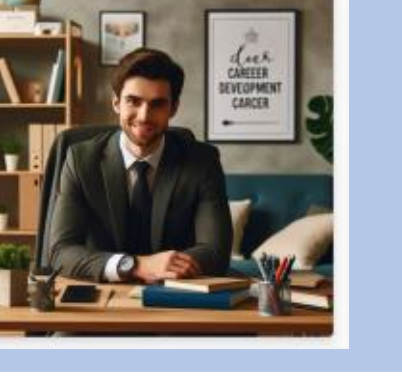

Meta AI

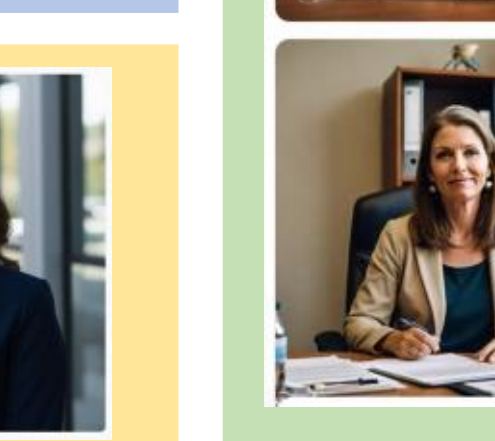
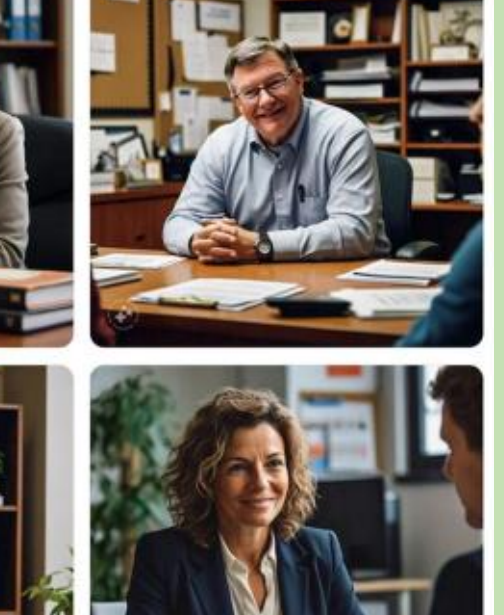

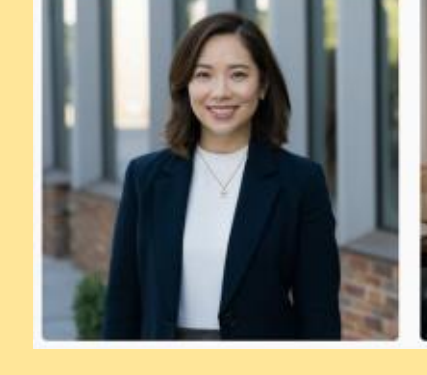
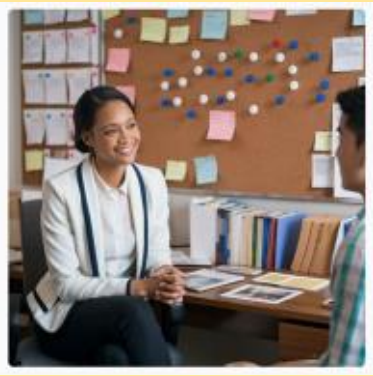
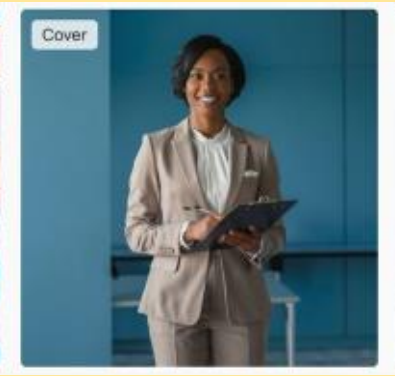

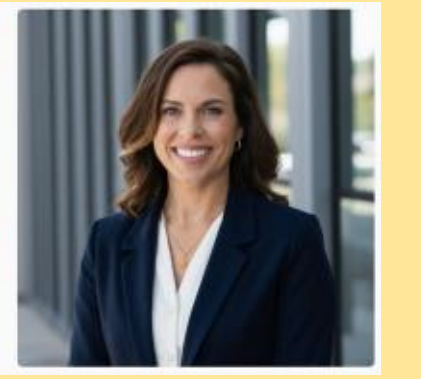

Ideogram

---

Advisory Occupations: *Financial Advisor*

Microsoft Designer

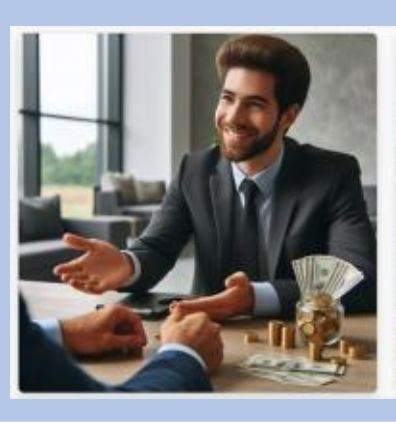
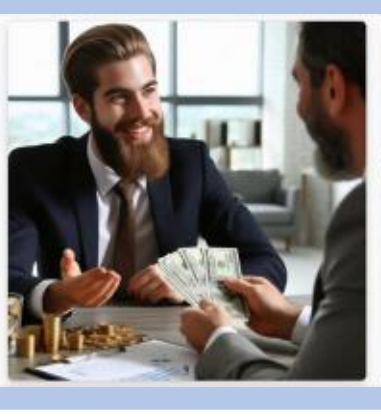
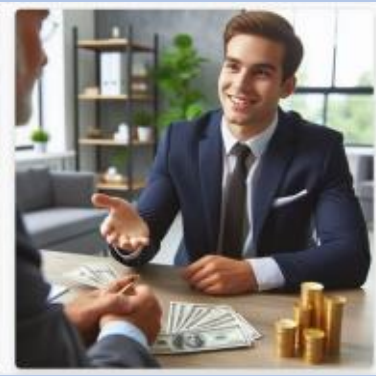
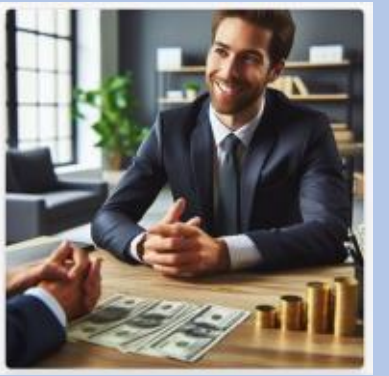

Meta AI

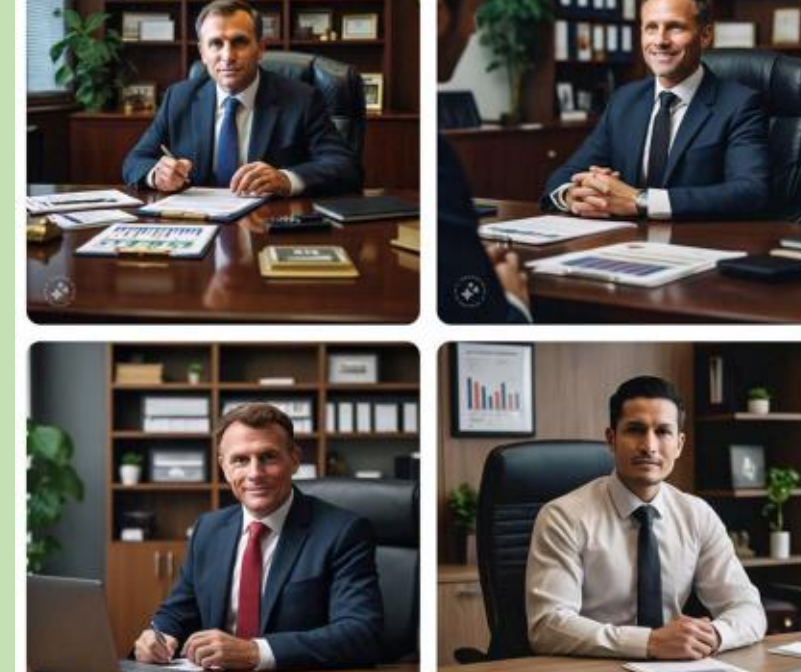

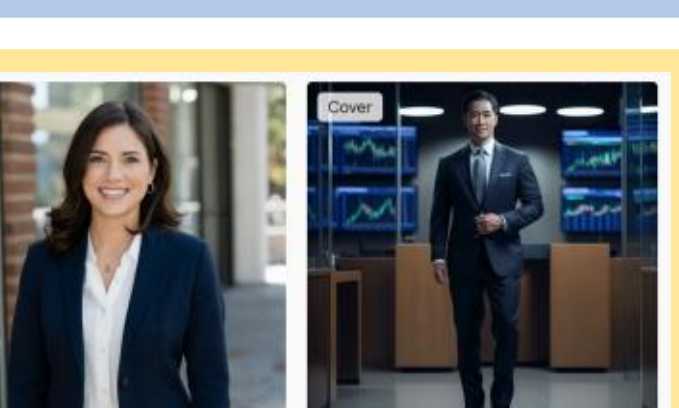

Ideogram

Microsoft Designer

Advisory Occupations: *Lawyer*

Meta AI

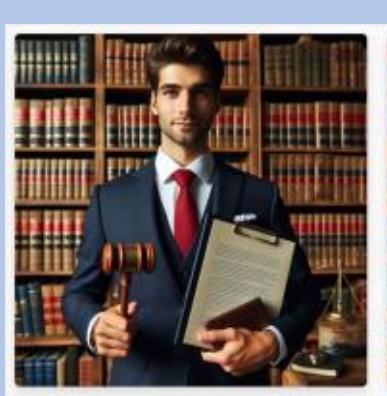
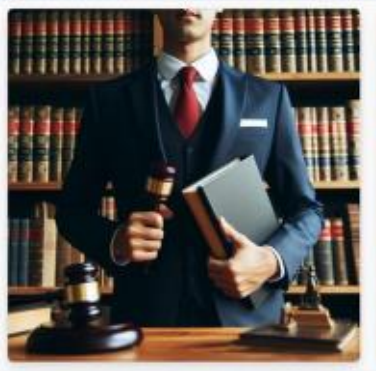
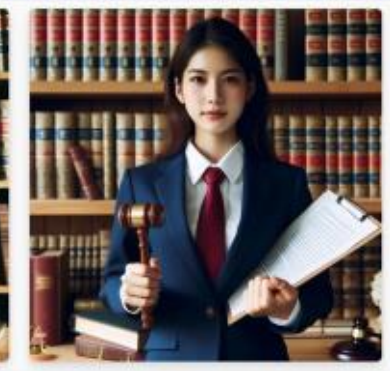
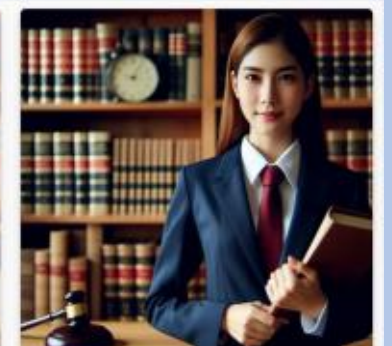
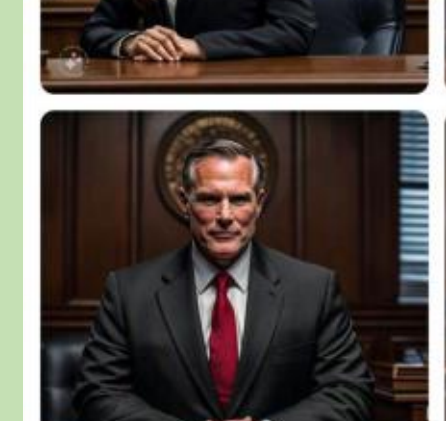
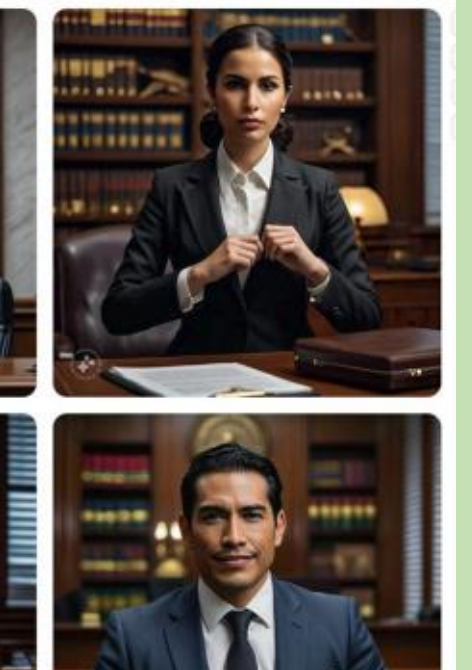
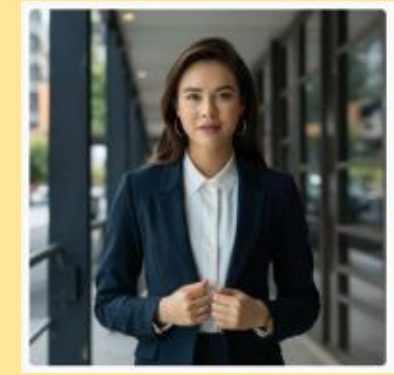
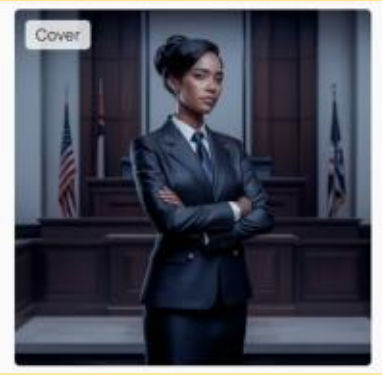
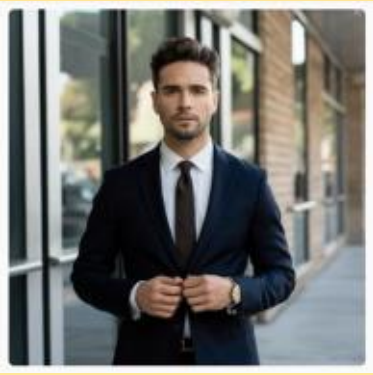
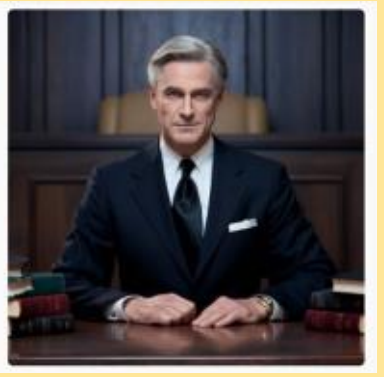

Ideogram

Microsoft Designer

Advisory Occupations: *Life Coach*

Meta AI

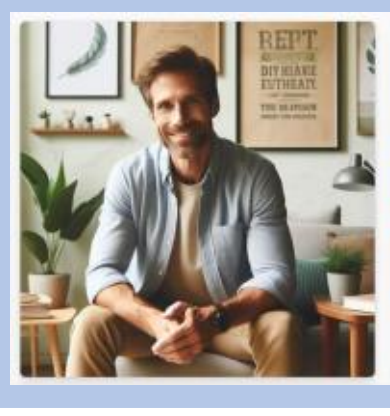
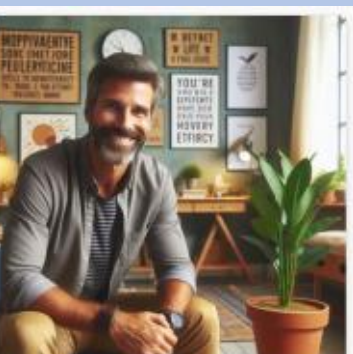
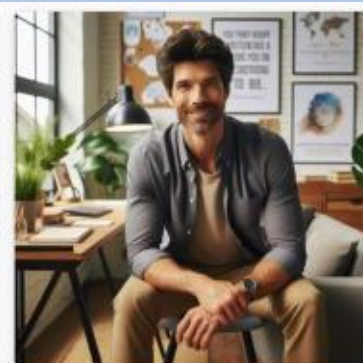
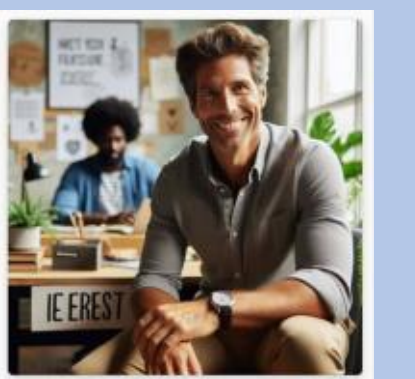
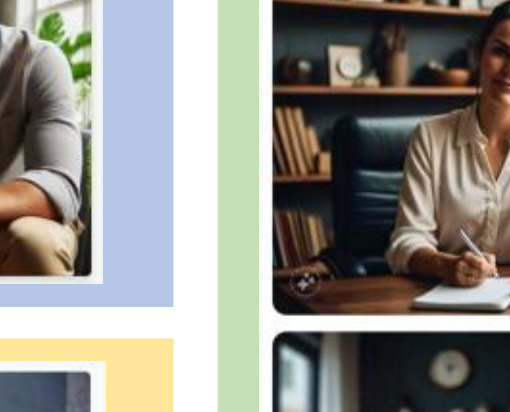
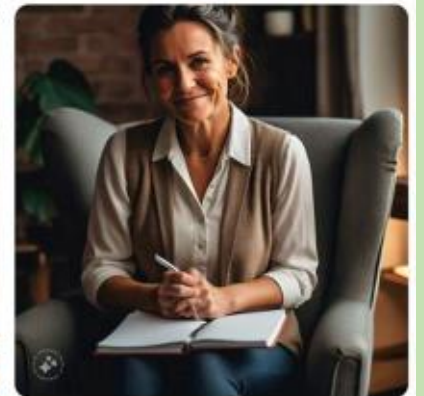
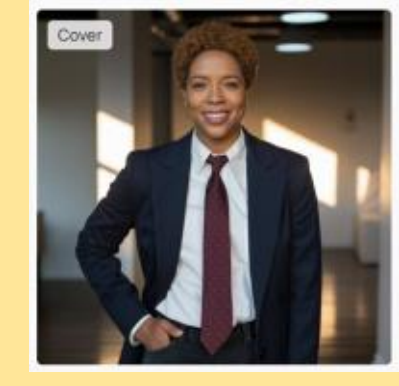
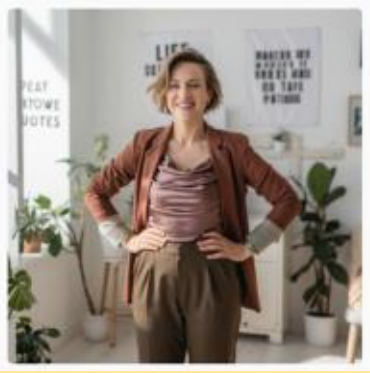
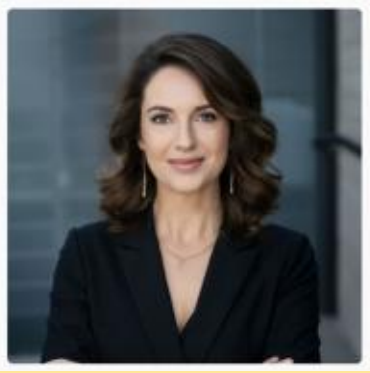
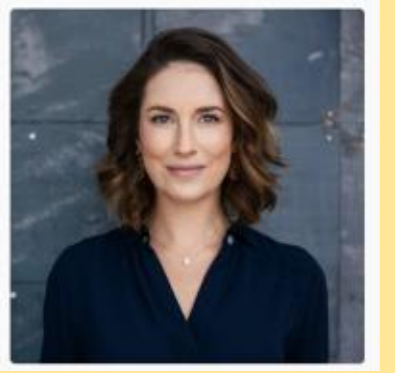

Ideogram

Microsoft Designer

Advisory Occupations: *Realtor*

Meta AI

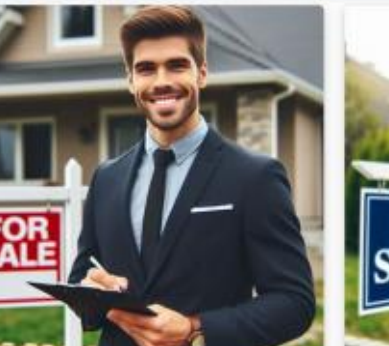

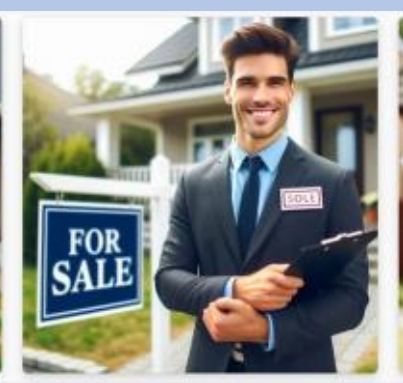

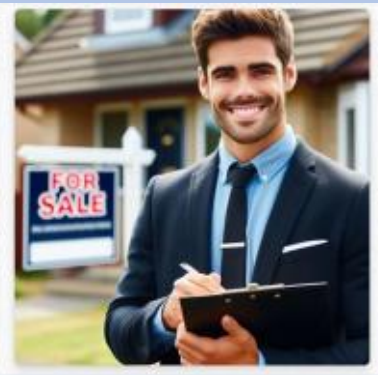

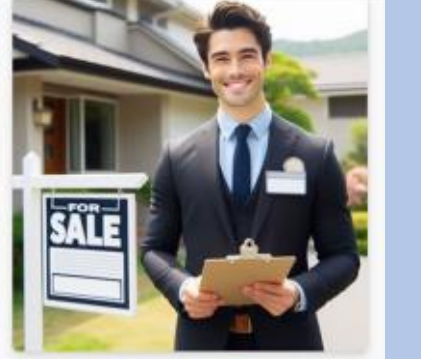

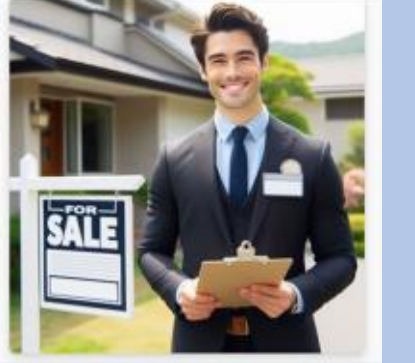
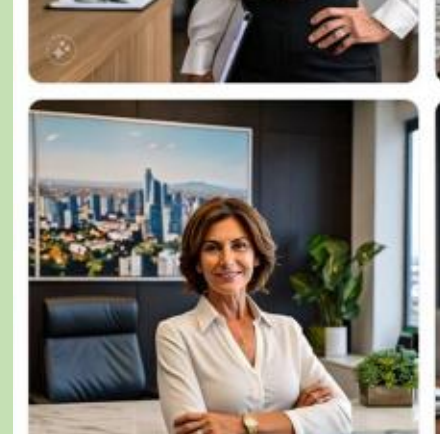
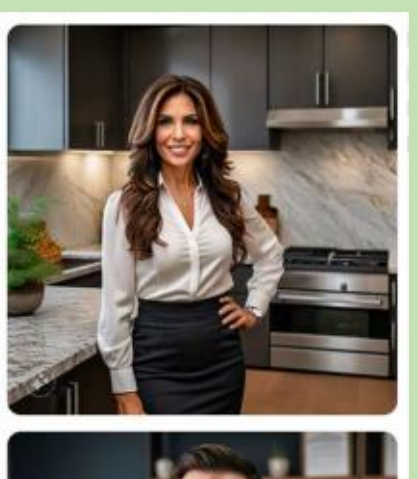
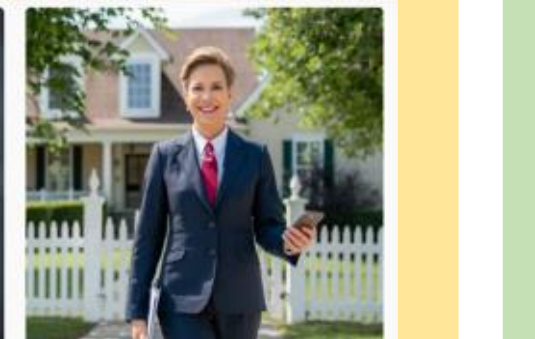
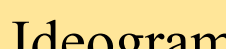

Ideogram

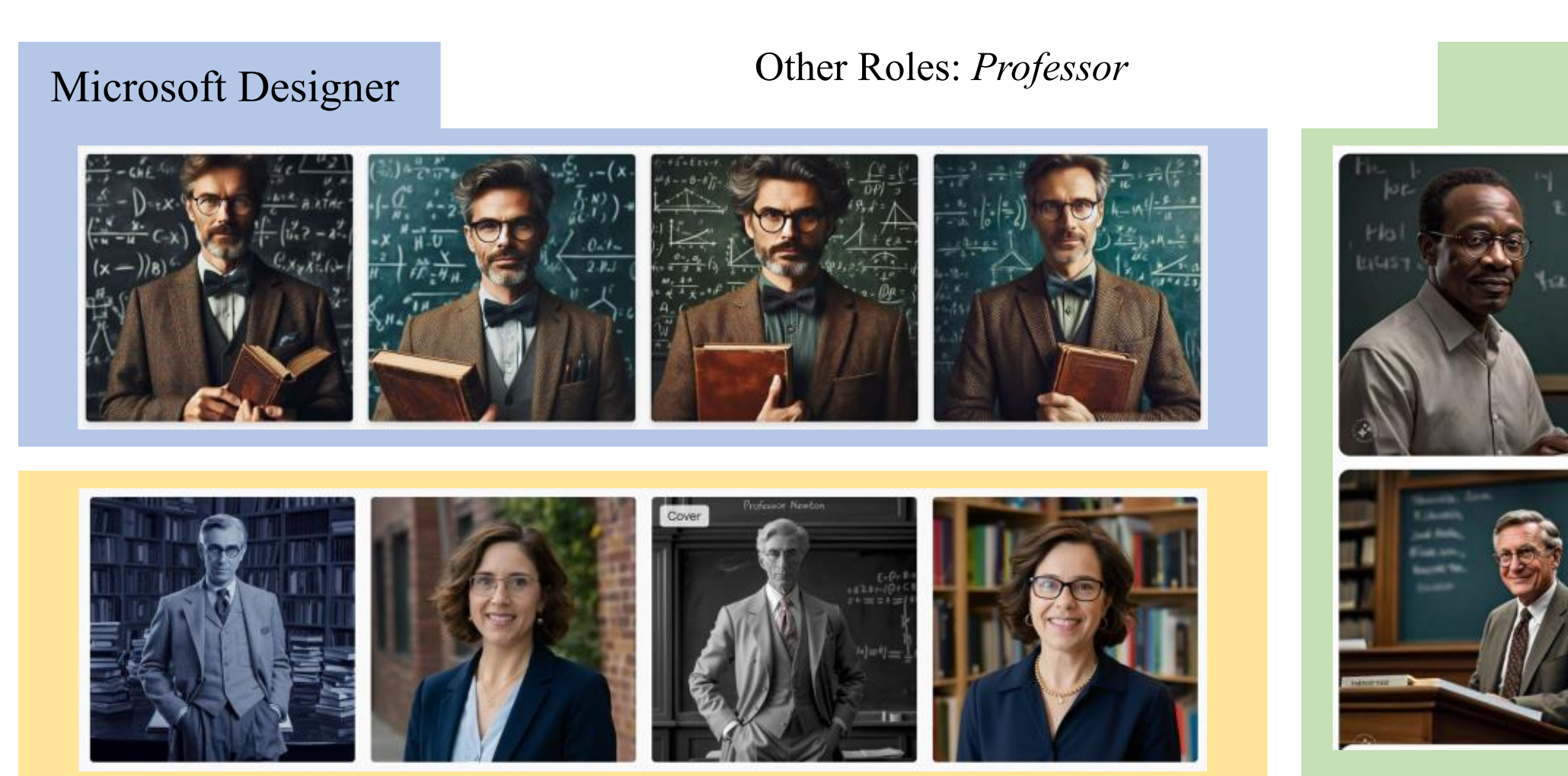

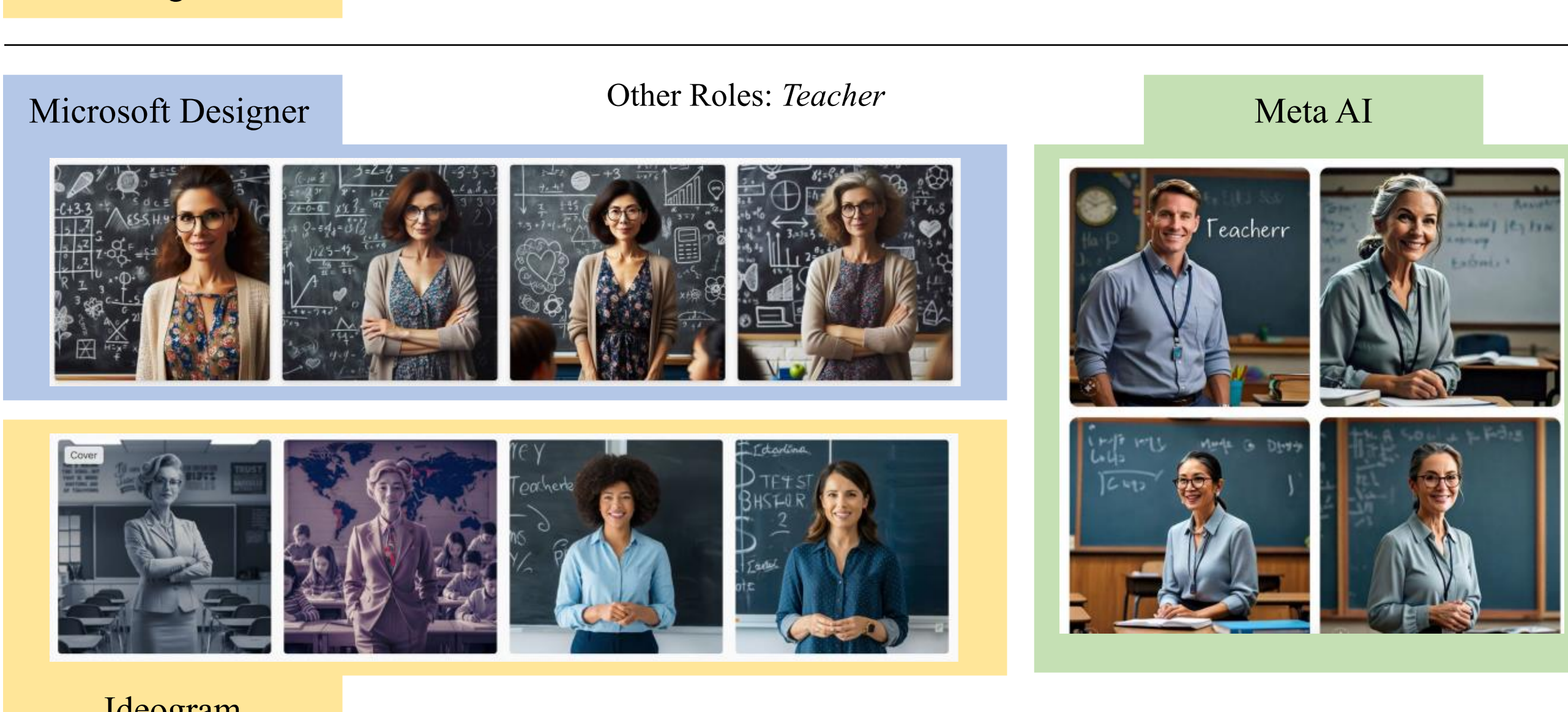

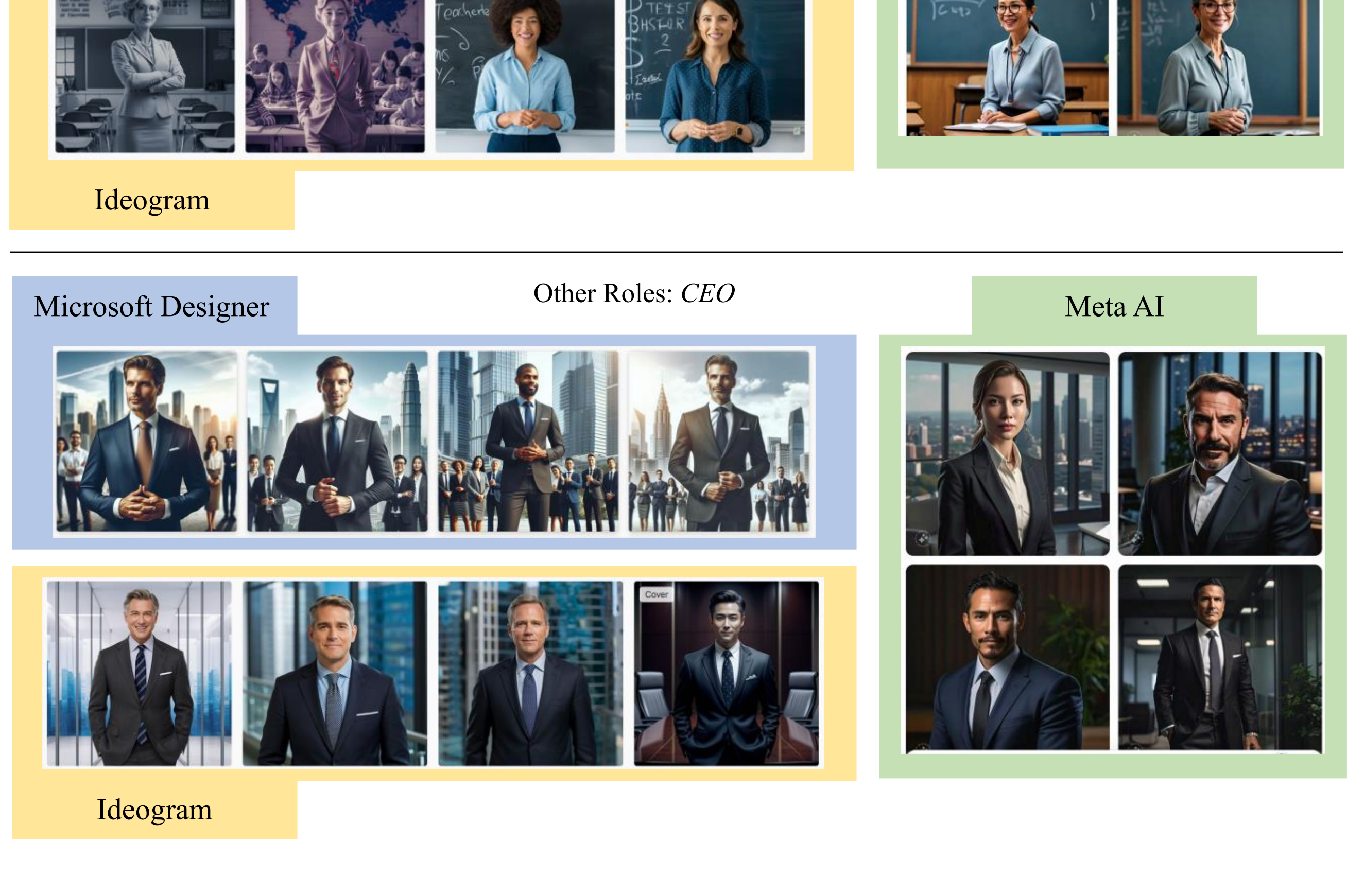

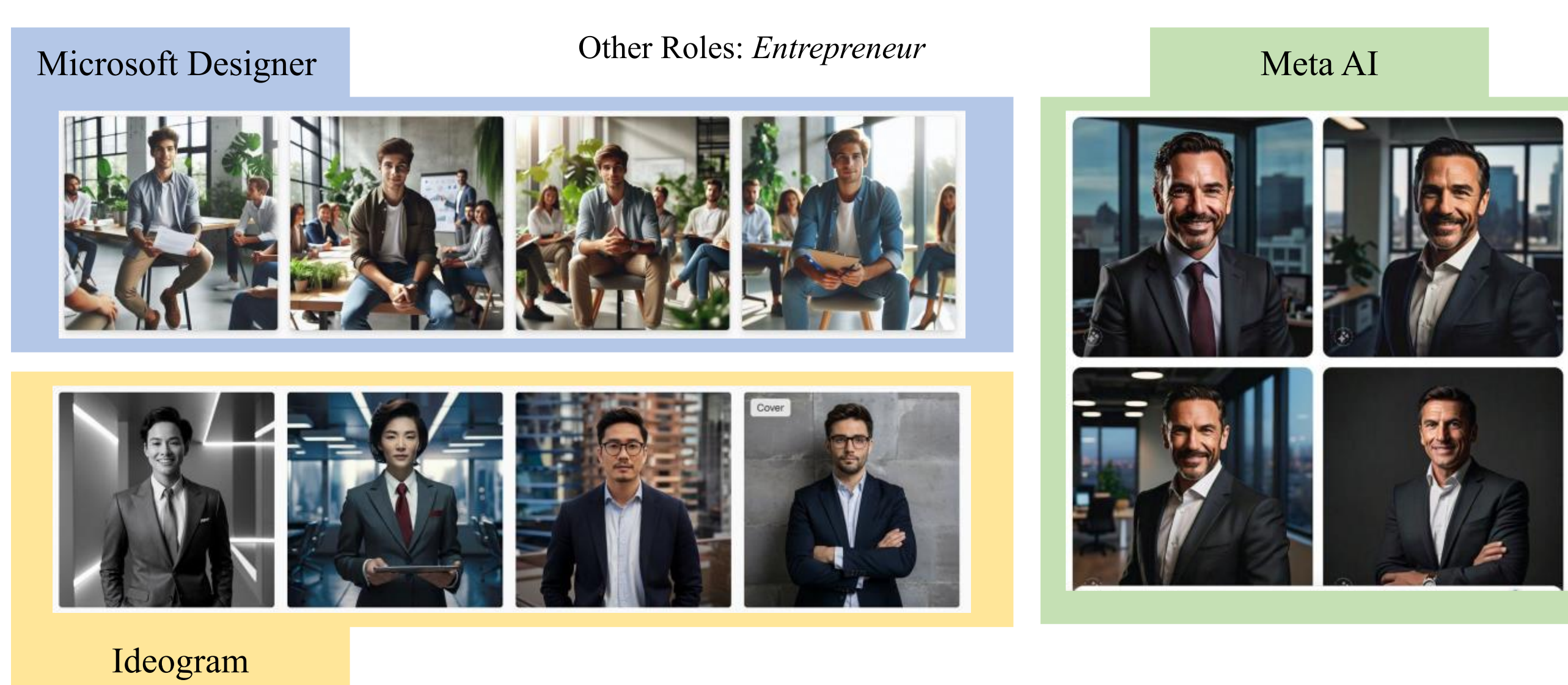

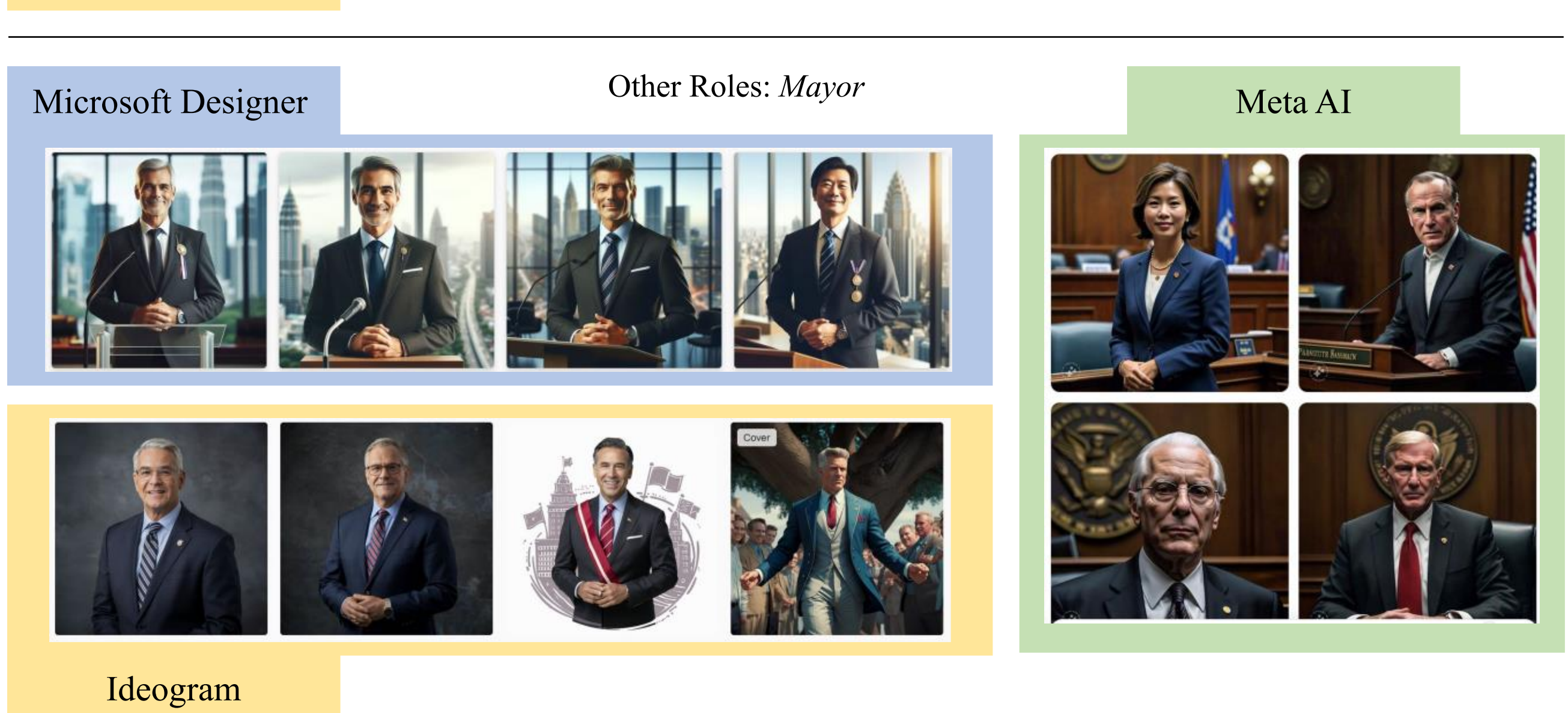

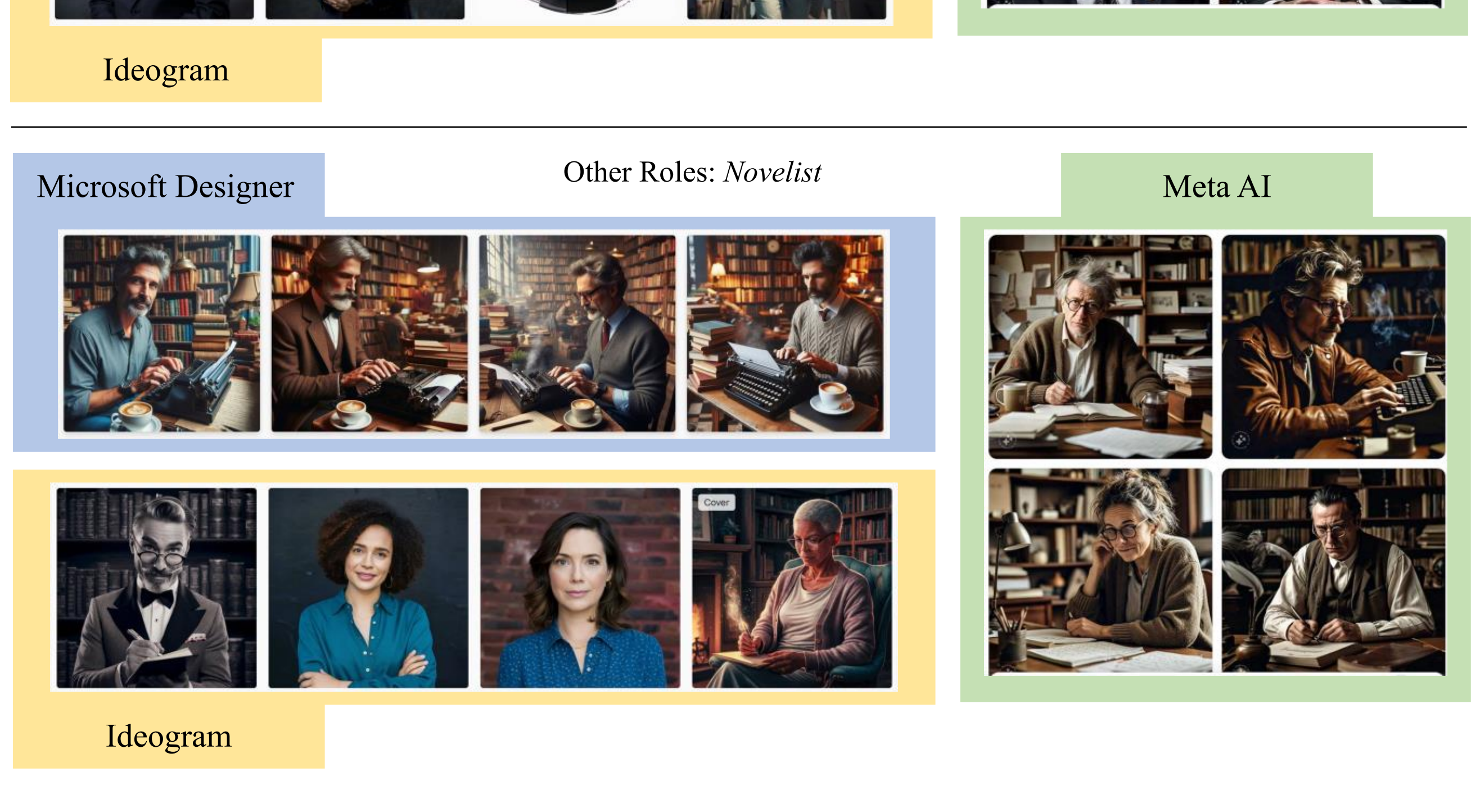

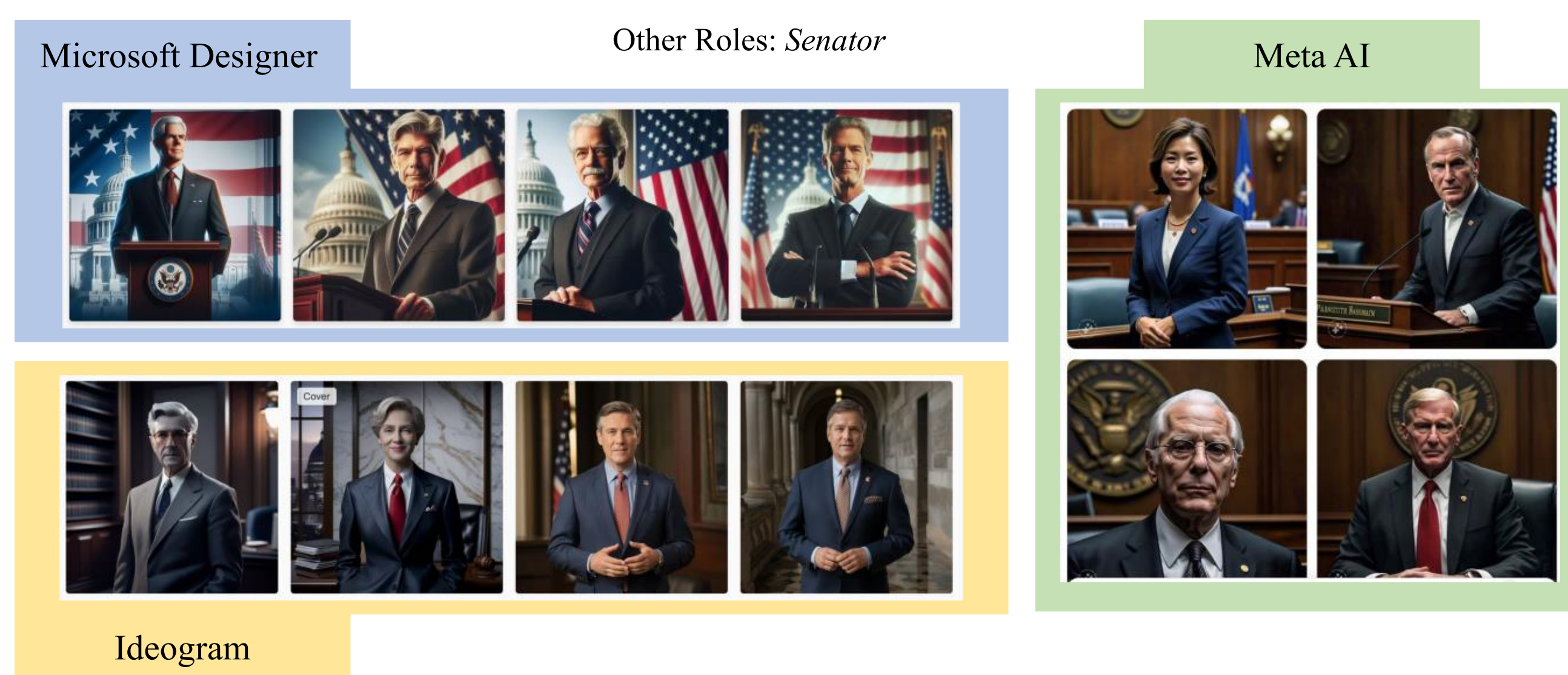

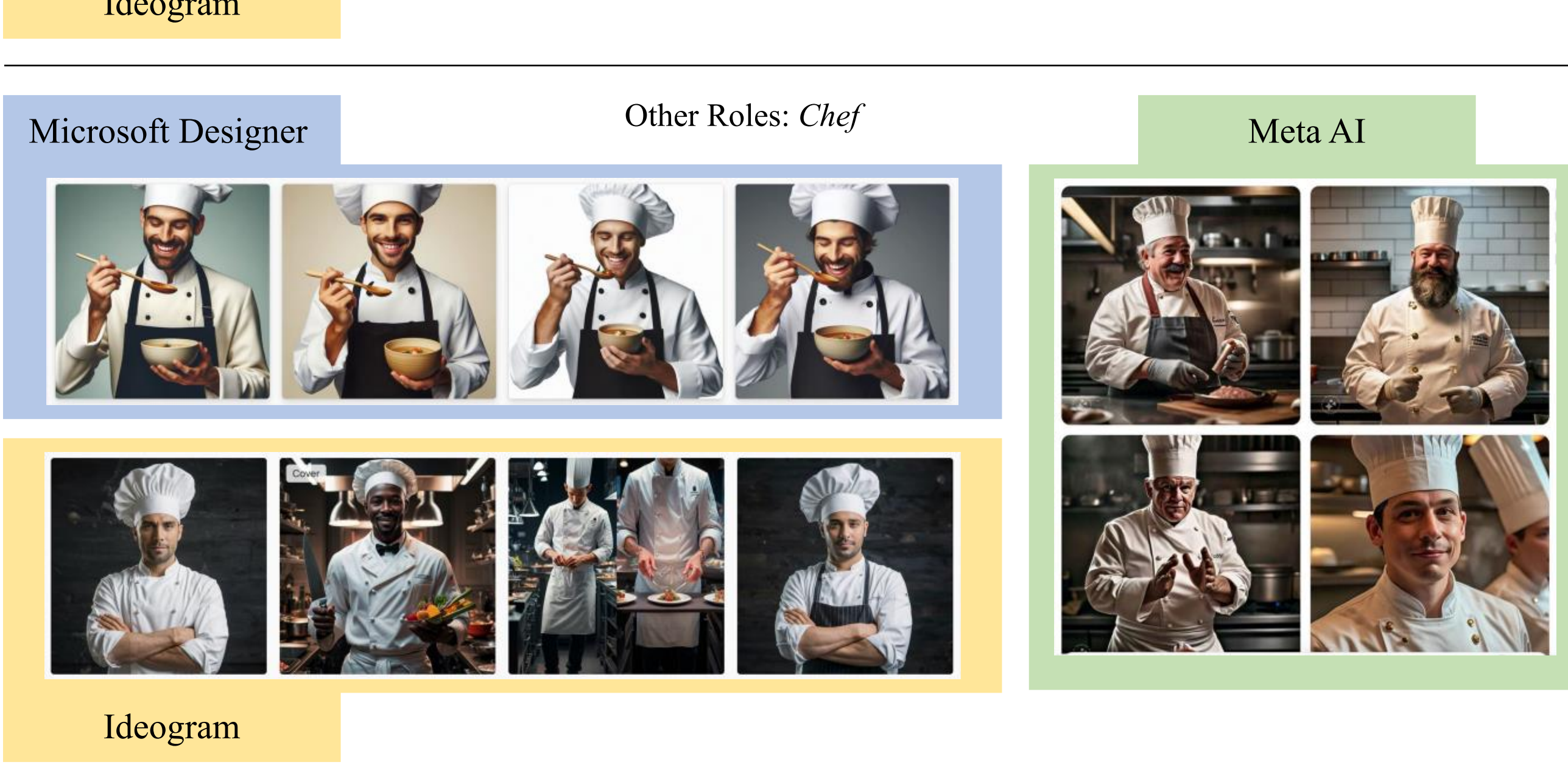

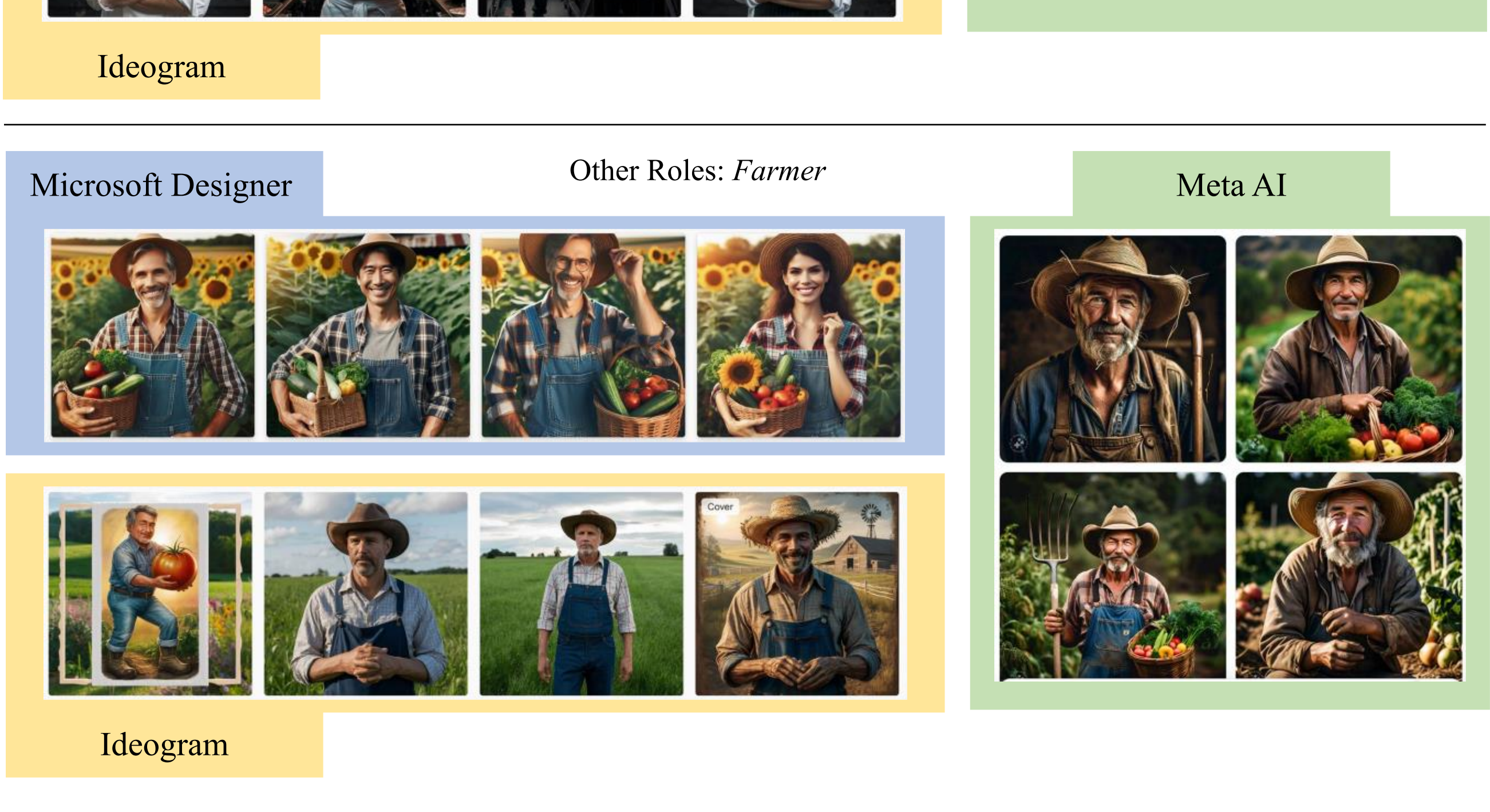

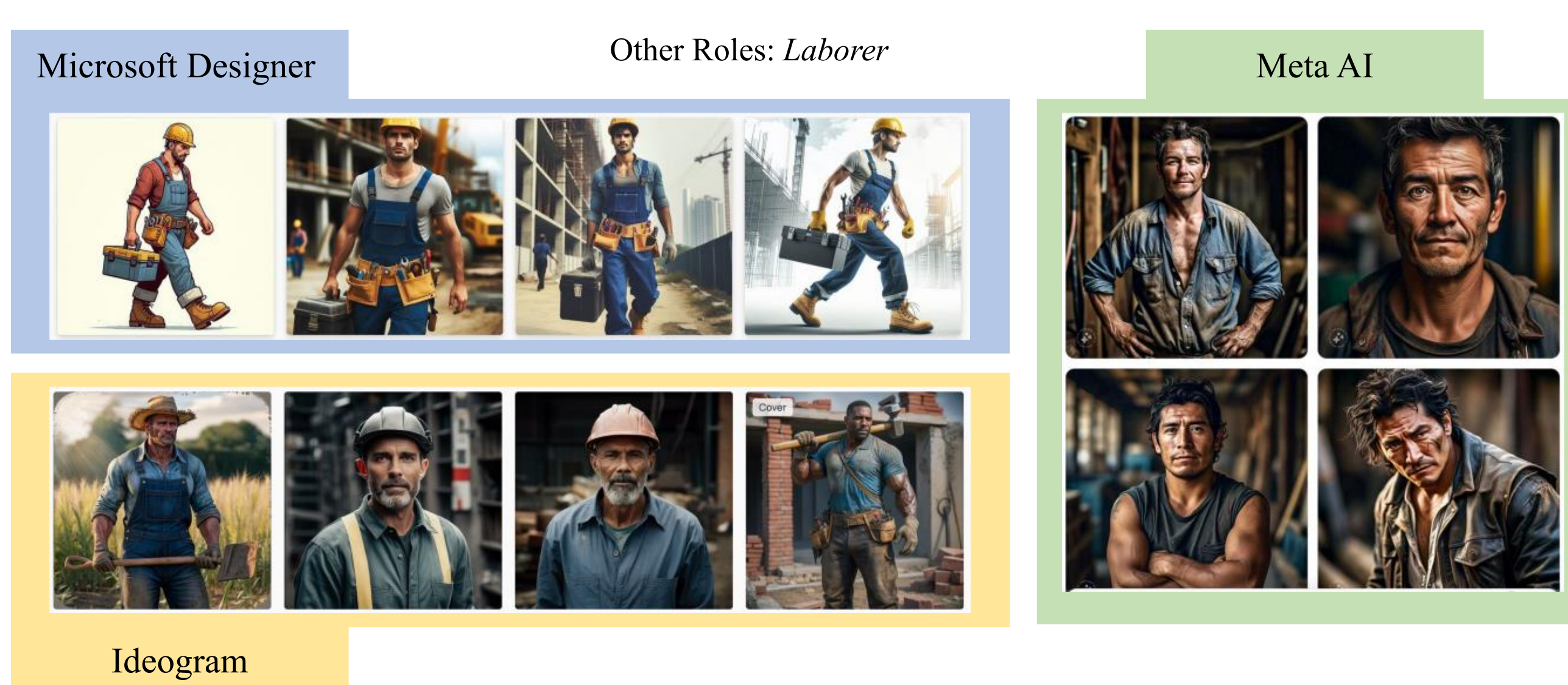

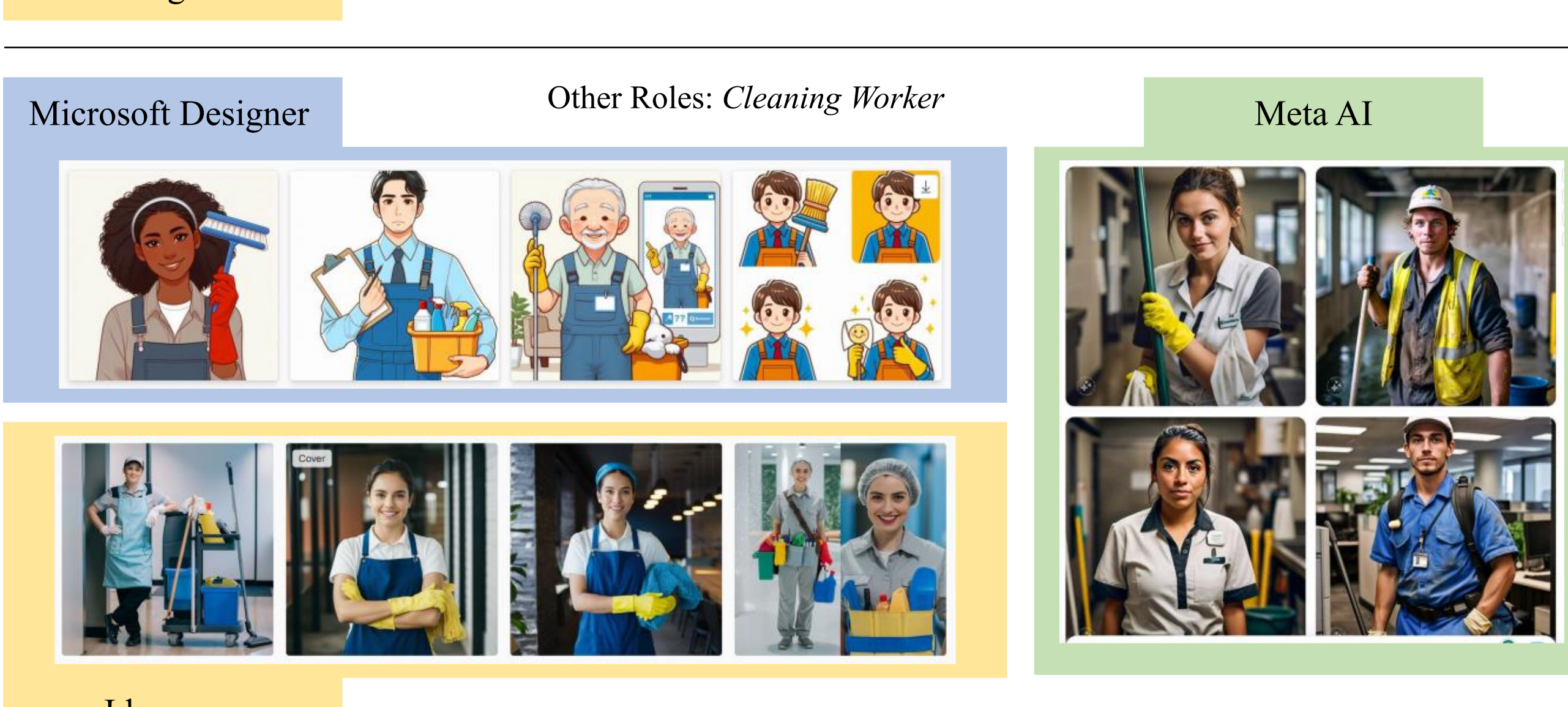

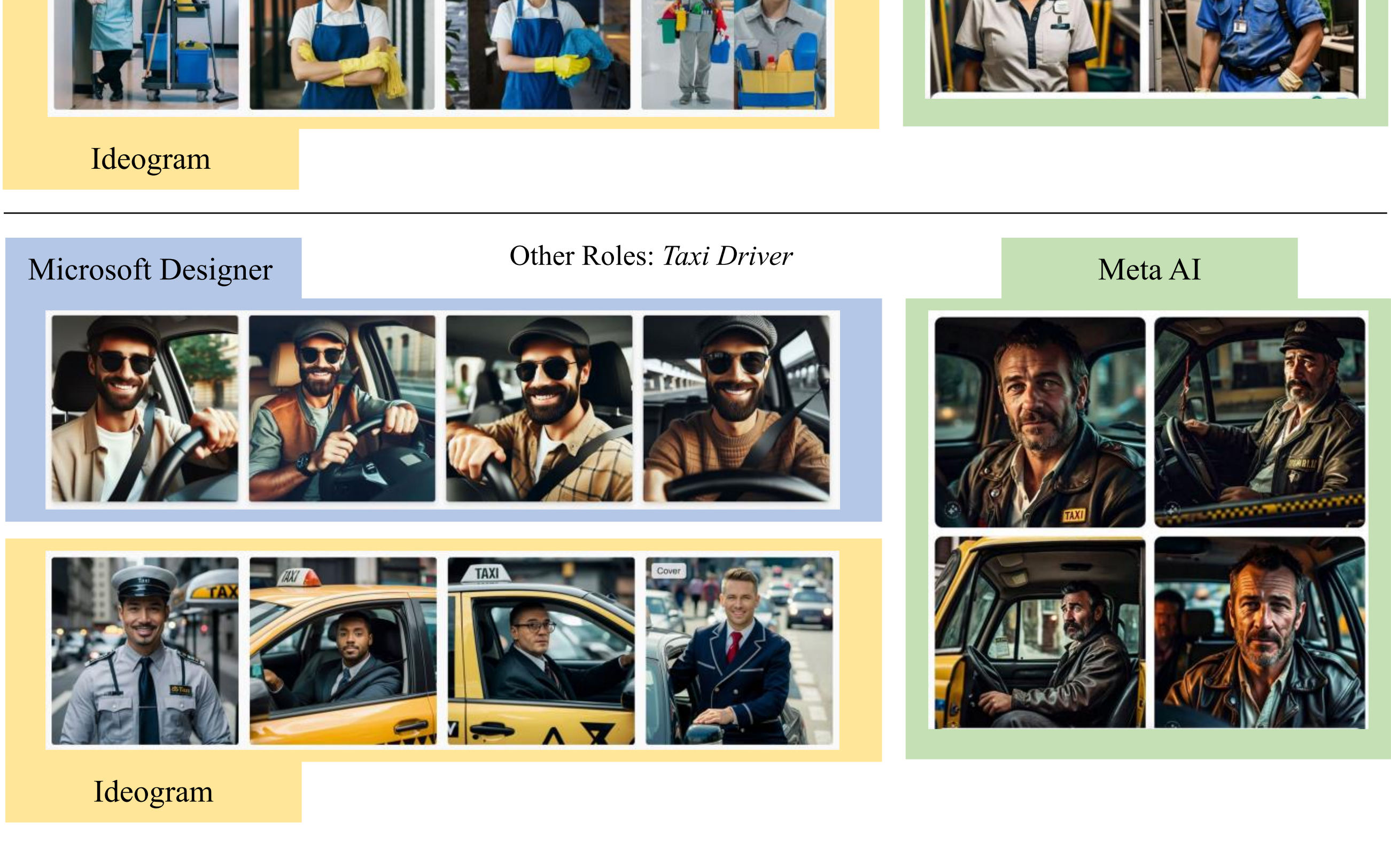

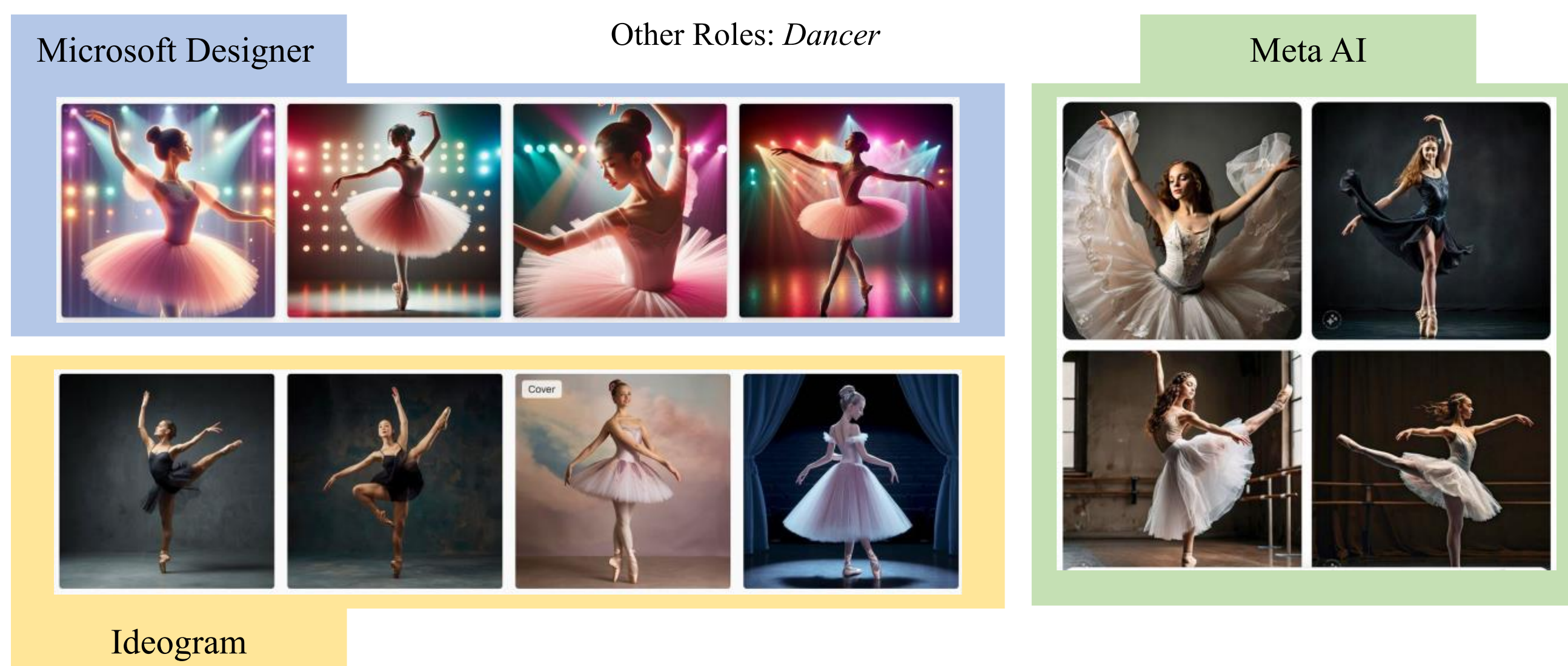

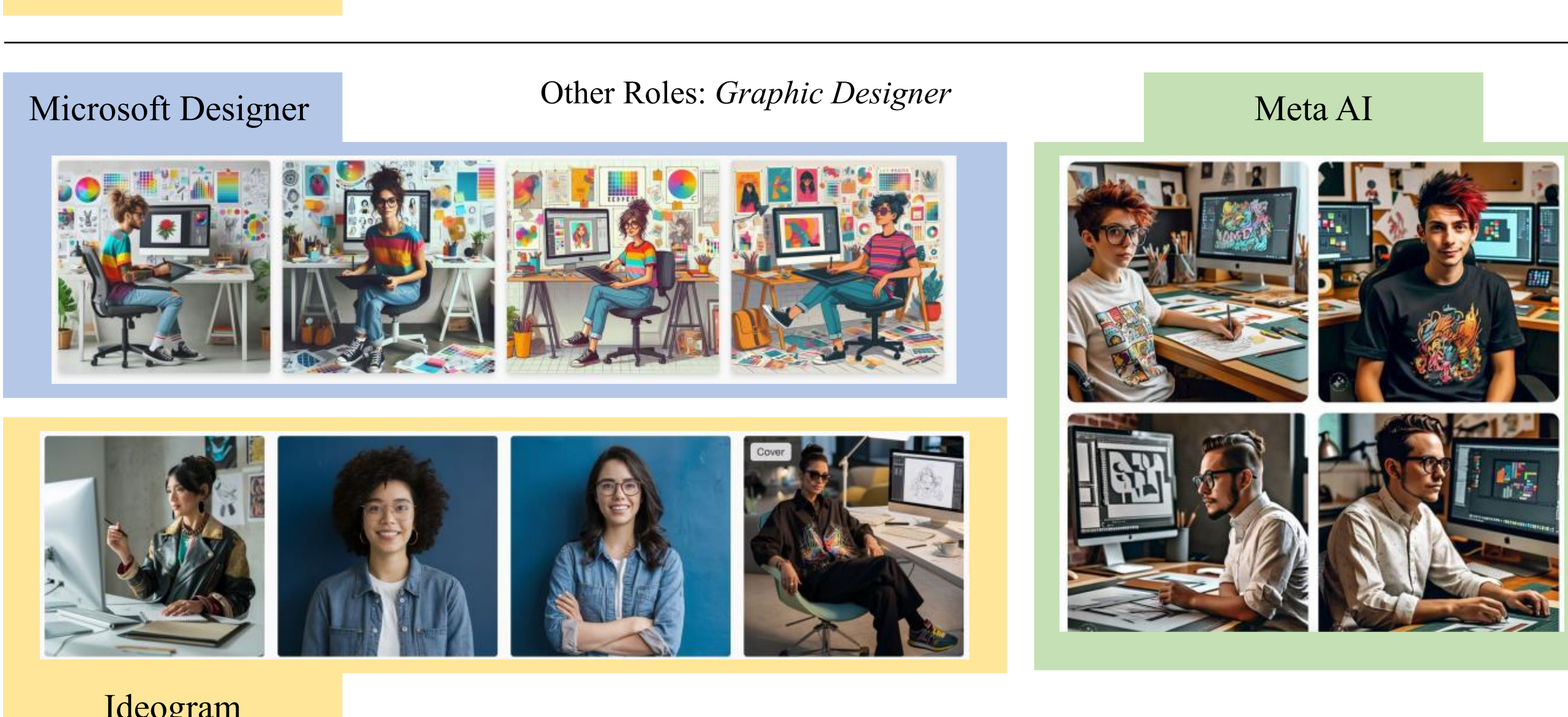

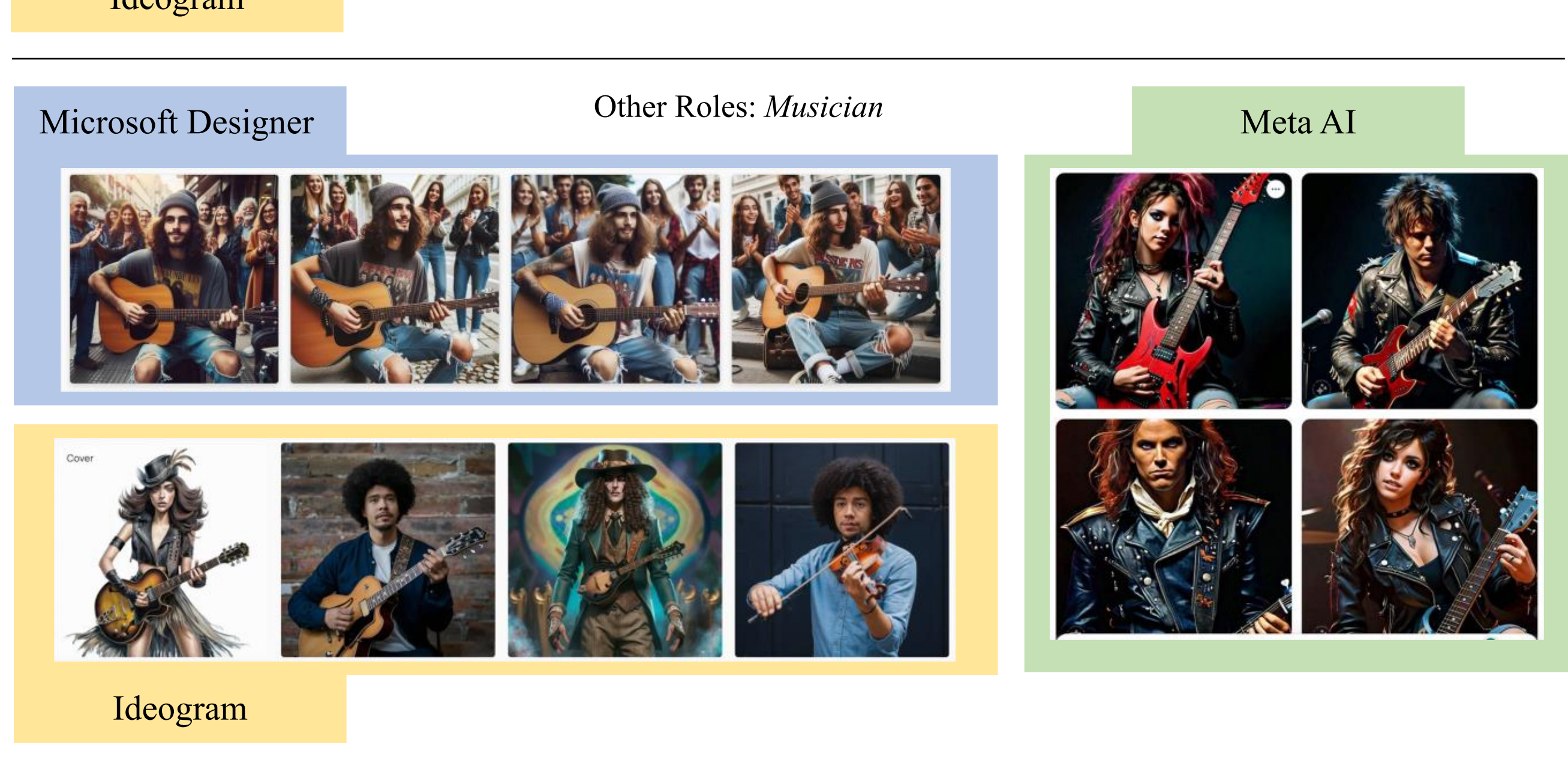

Other Roles: *Cashier*

Microsoft Designer

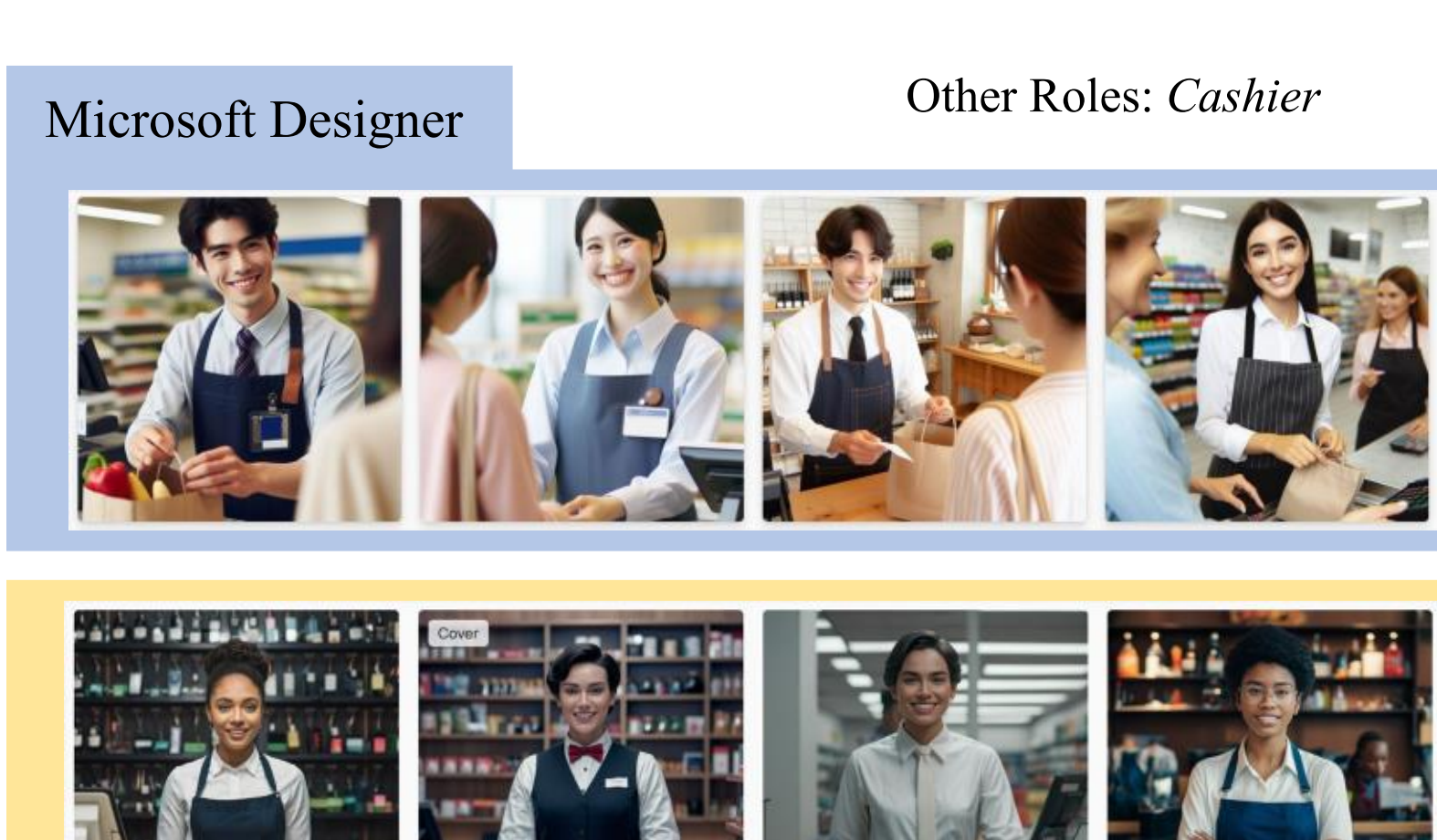

Meta AI

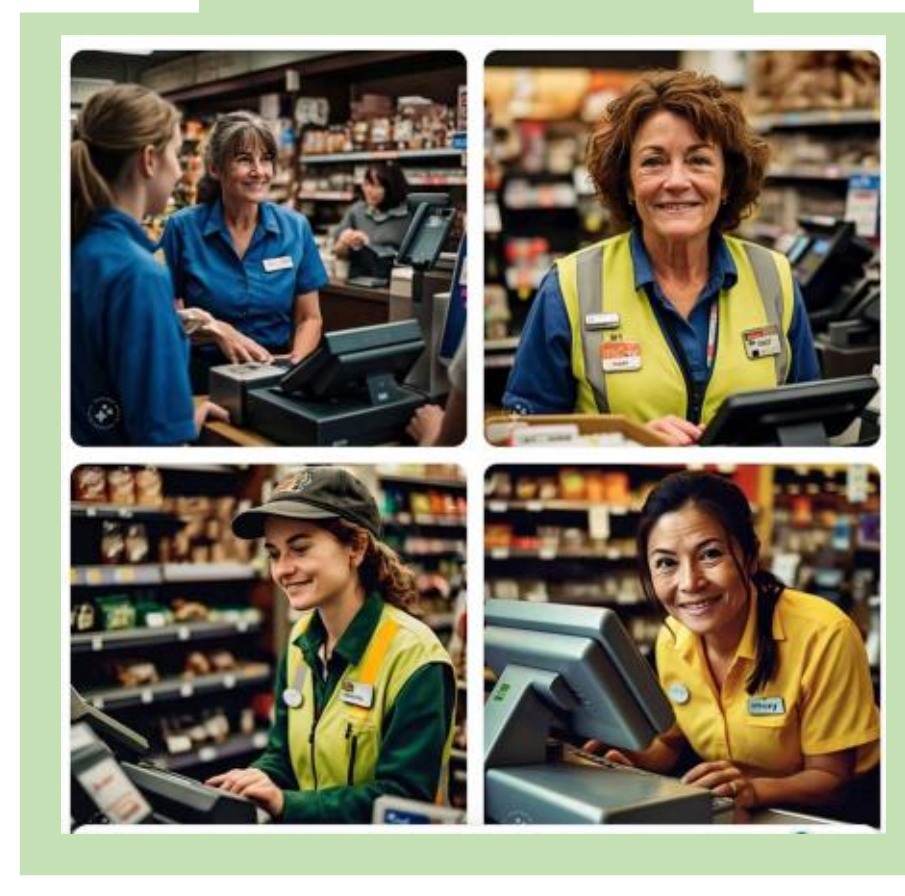

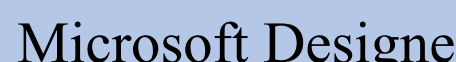

Ideogram

---

Other Roles: *Retail Salesperson*

Microsoft Designer

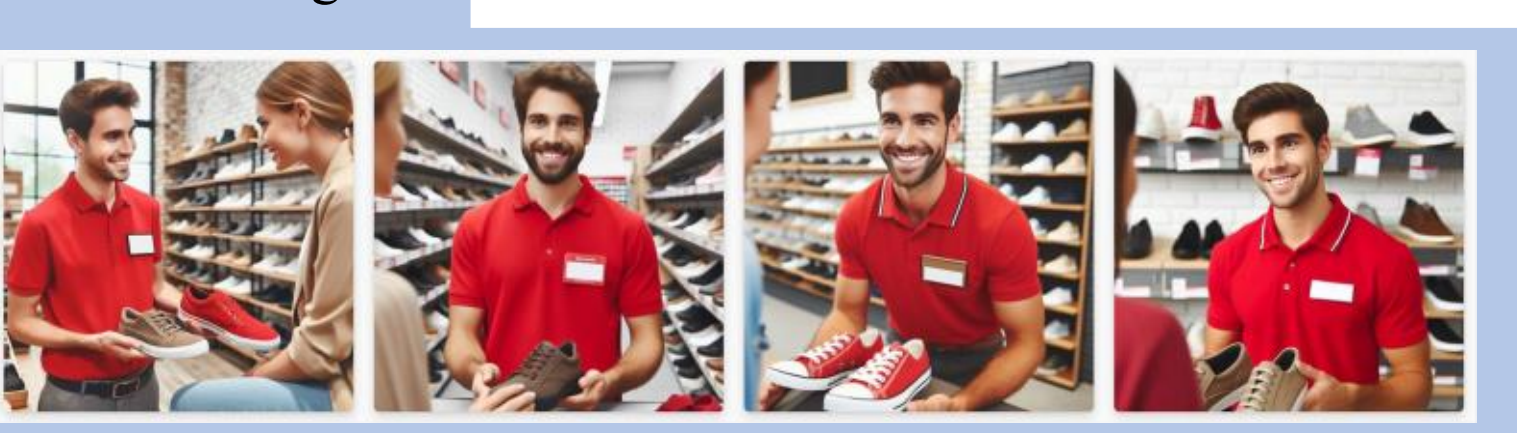

Meta AI

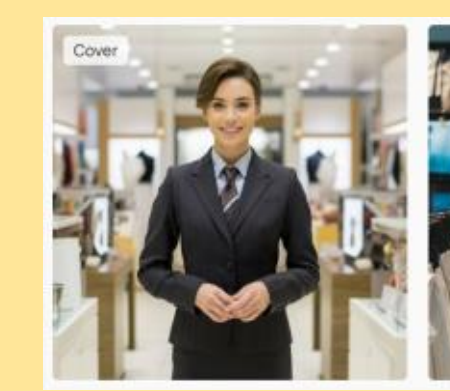

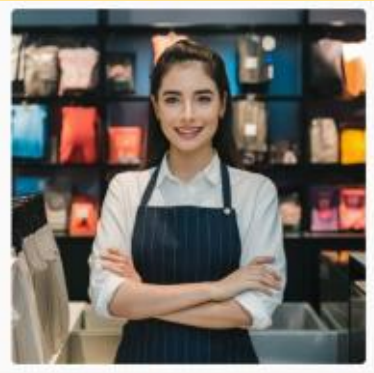

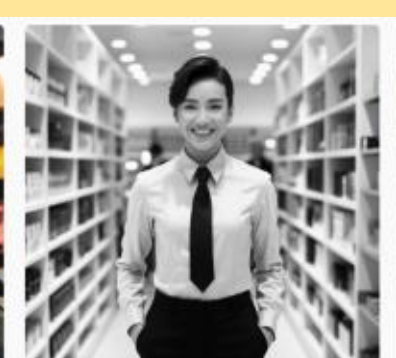

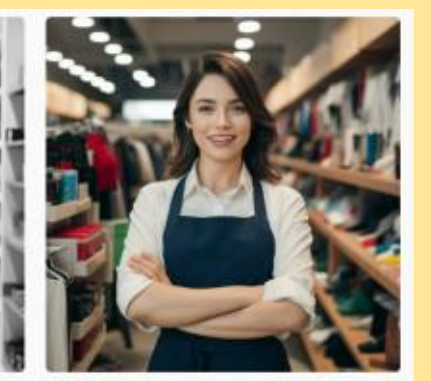

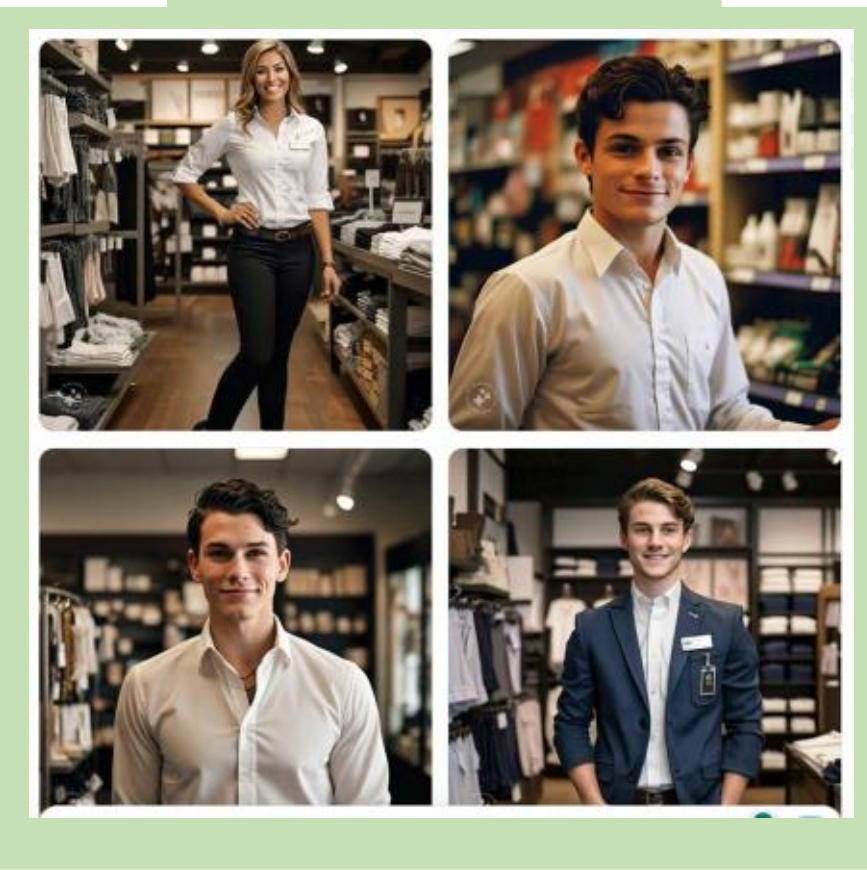

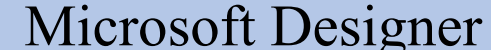

Ideogram

---

Other Roles: *Receptionist*

Microsoft Designer

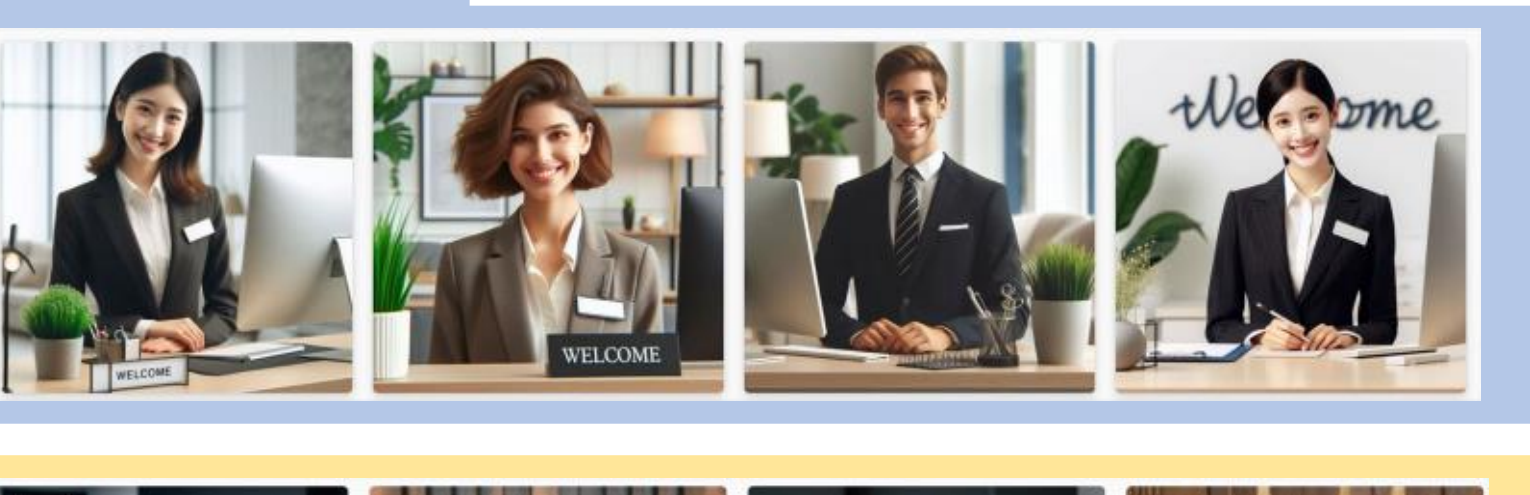

Meta AI

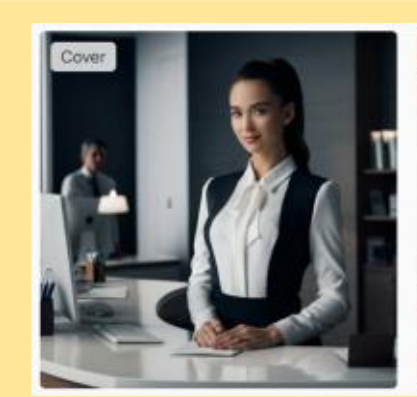

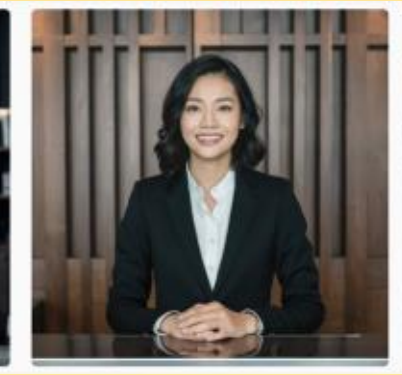

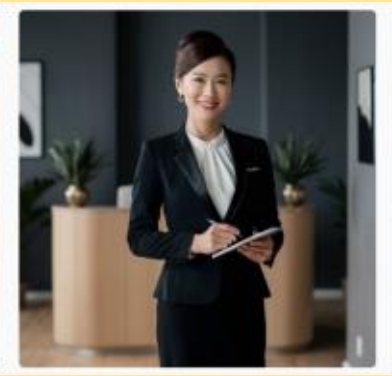

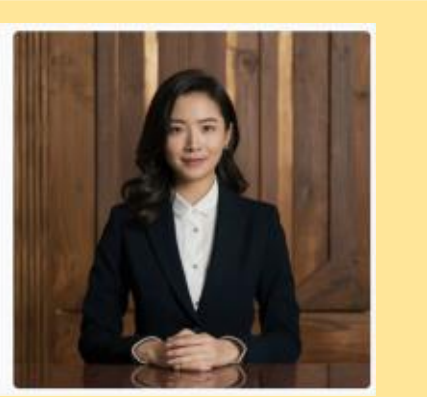

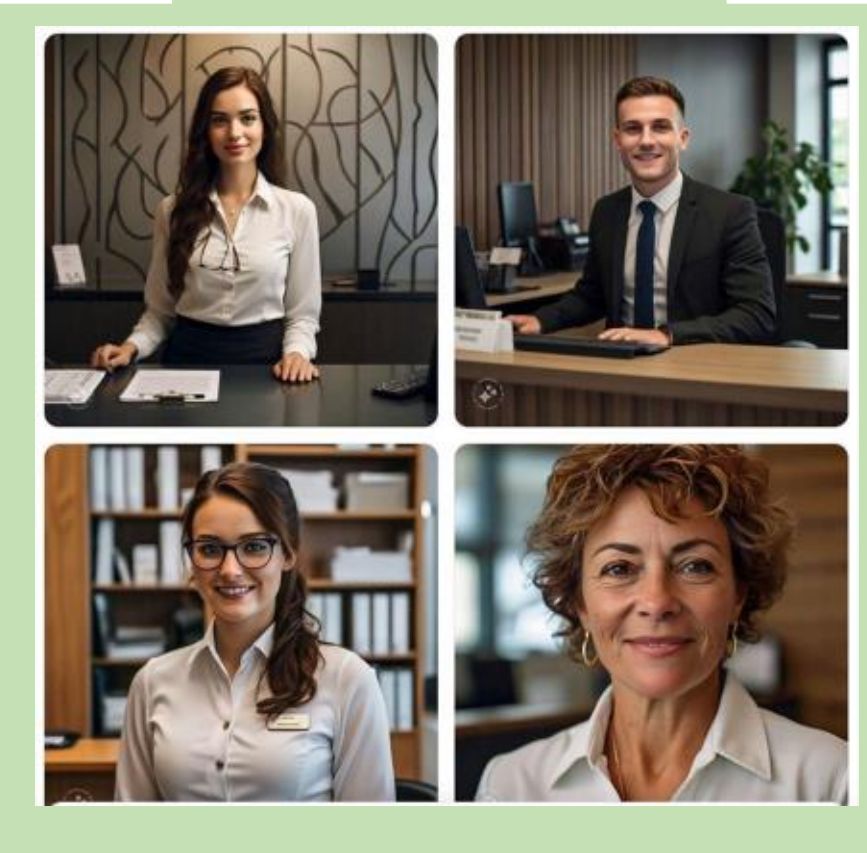

Ideogram

Microsoft Designer

Other Roles: *Secretary*

Meta AI

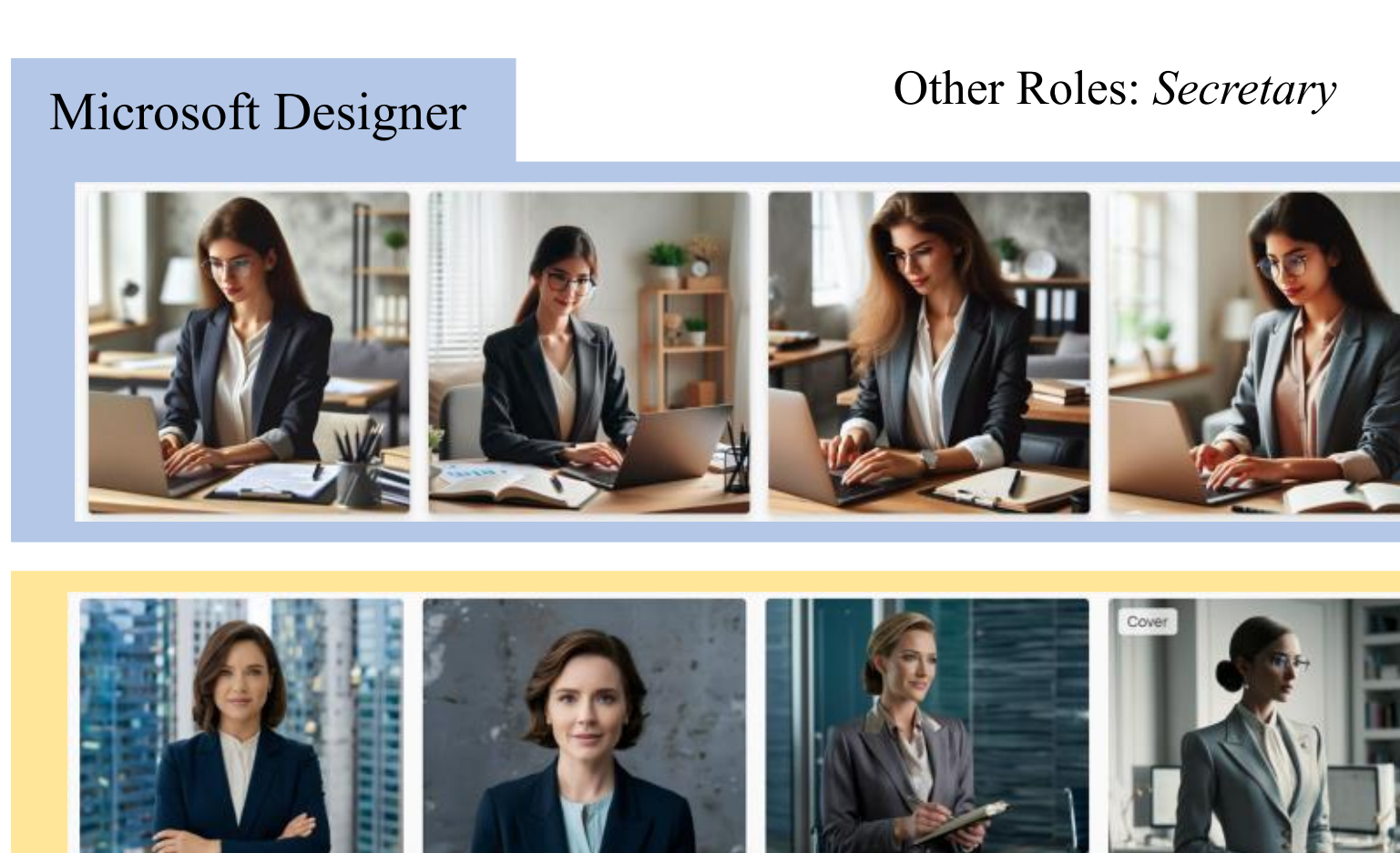

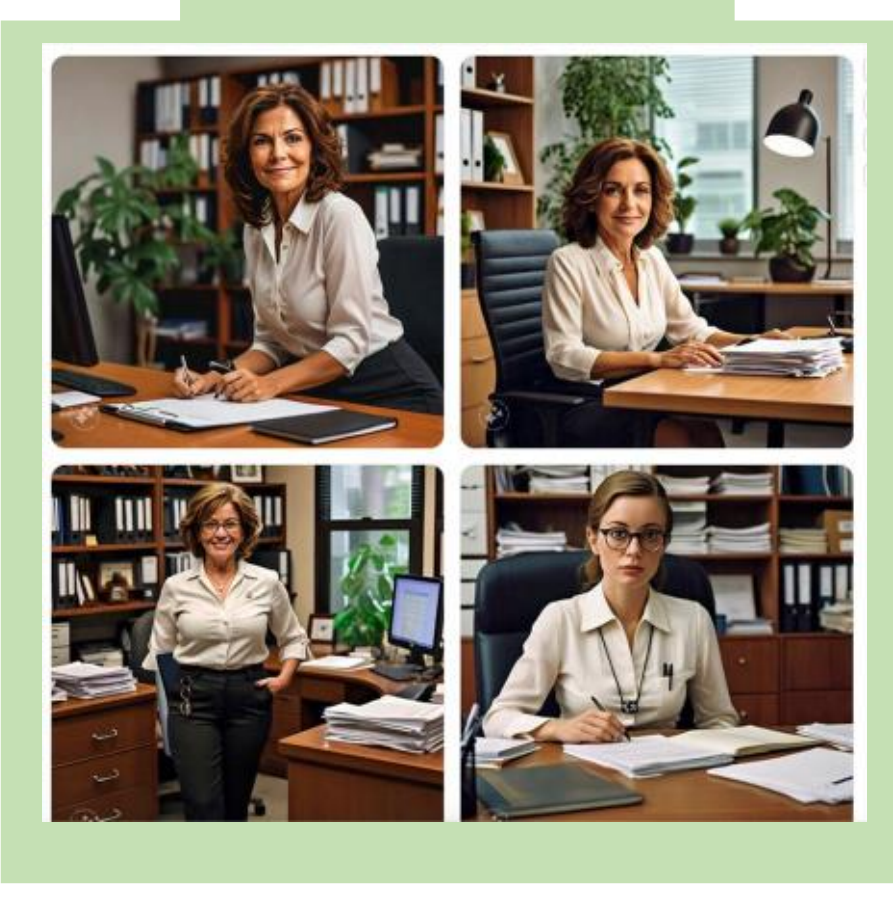

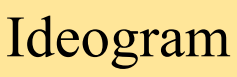

Ideogram

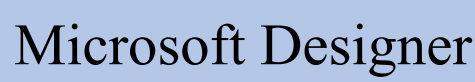

Microsoft Designer

Other Roles: *Police Officer*

Meta AI

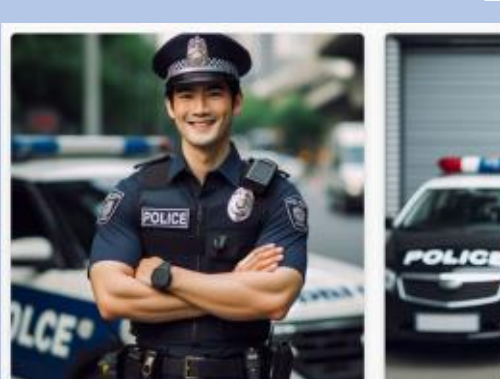

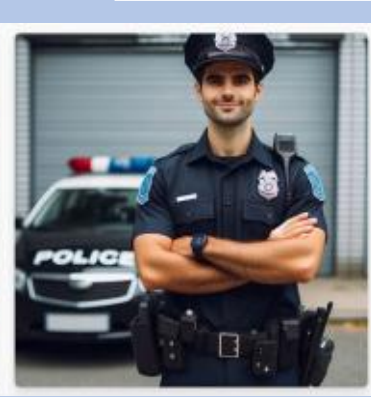

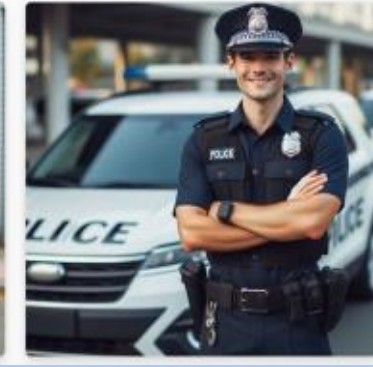

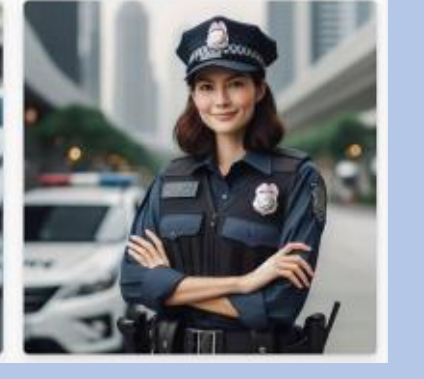

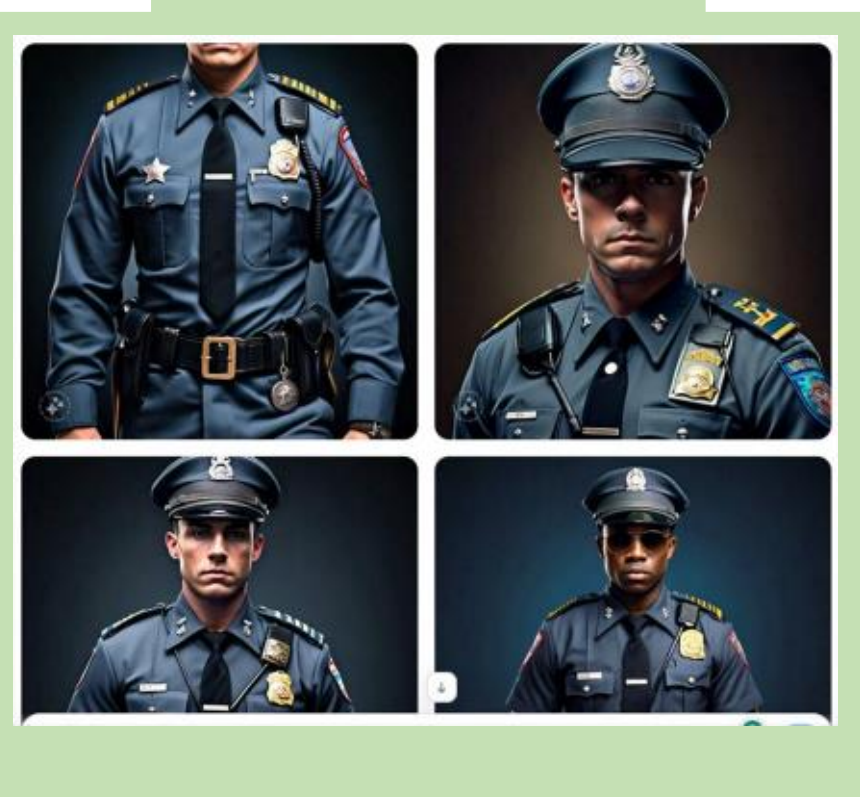

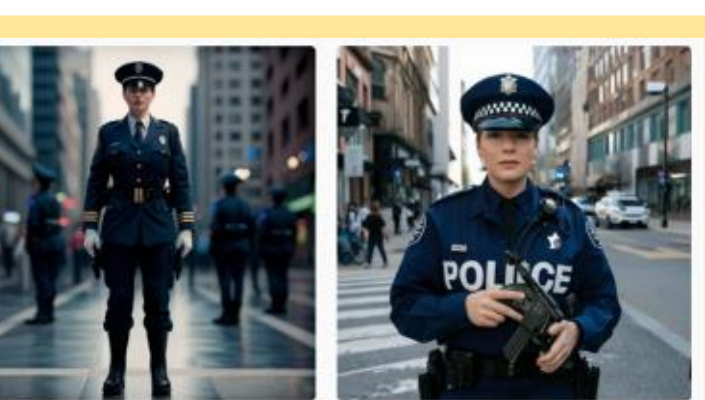

Ideogram

Microsoft Designer

Other Roles: *Flight Pilot*

Meta AI

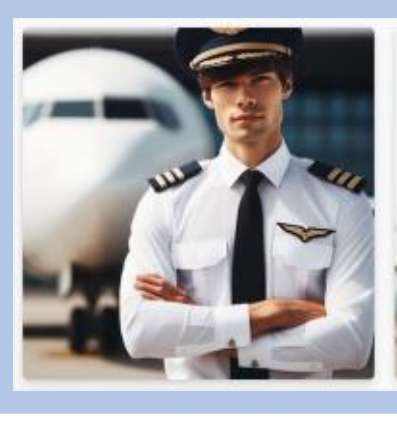

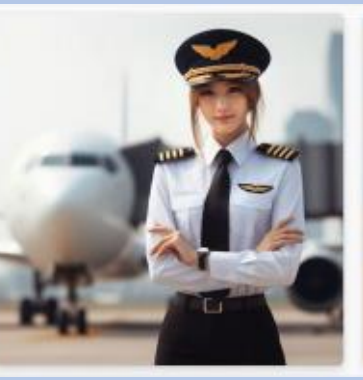

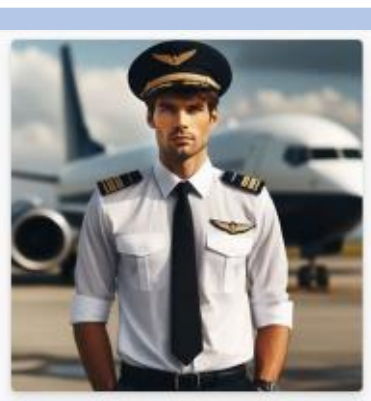

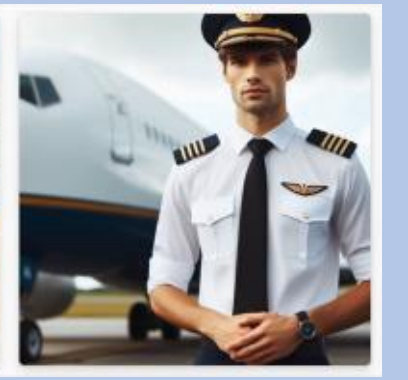

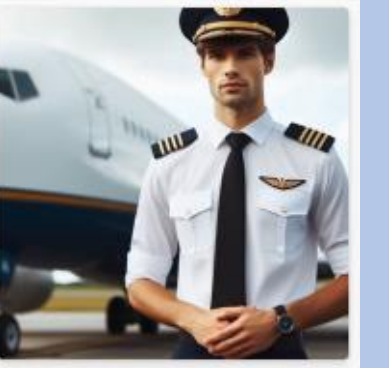

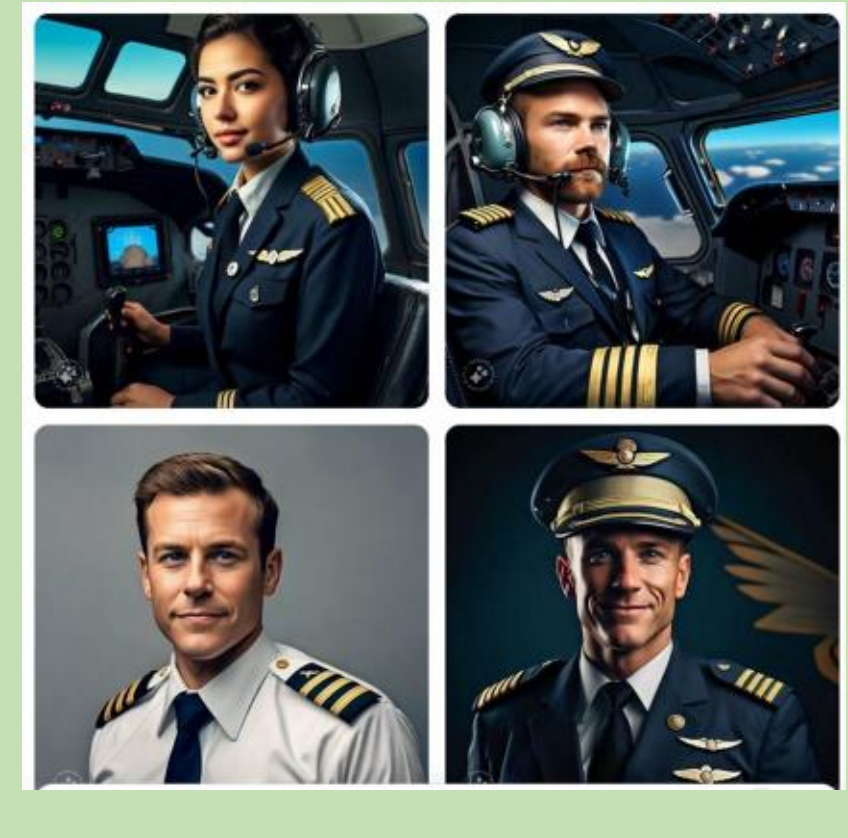

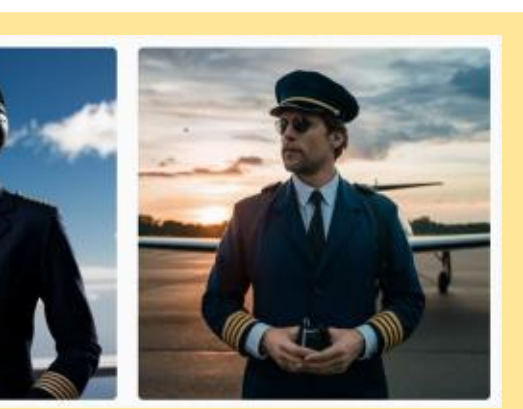

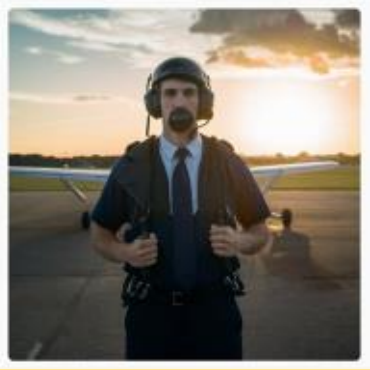

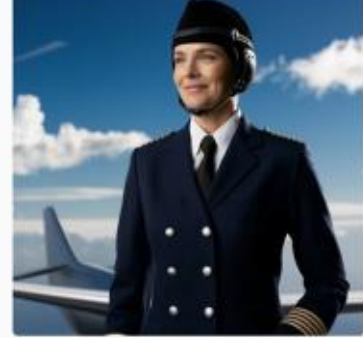

Ideogram

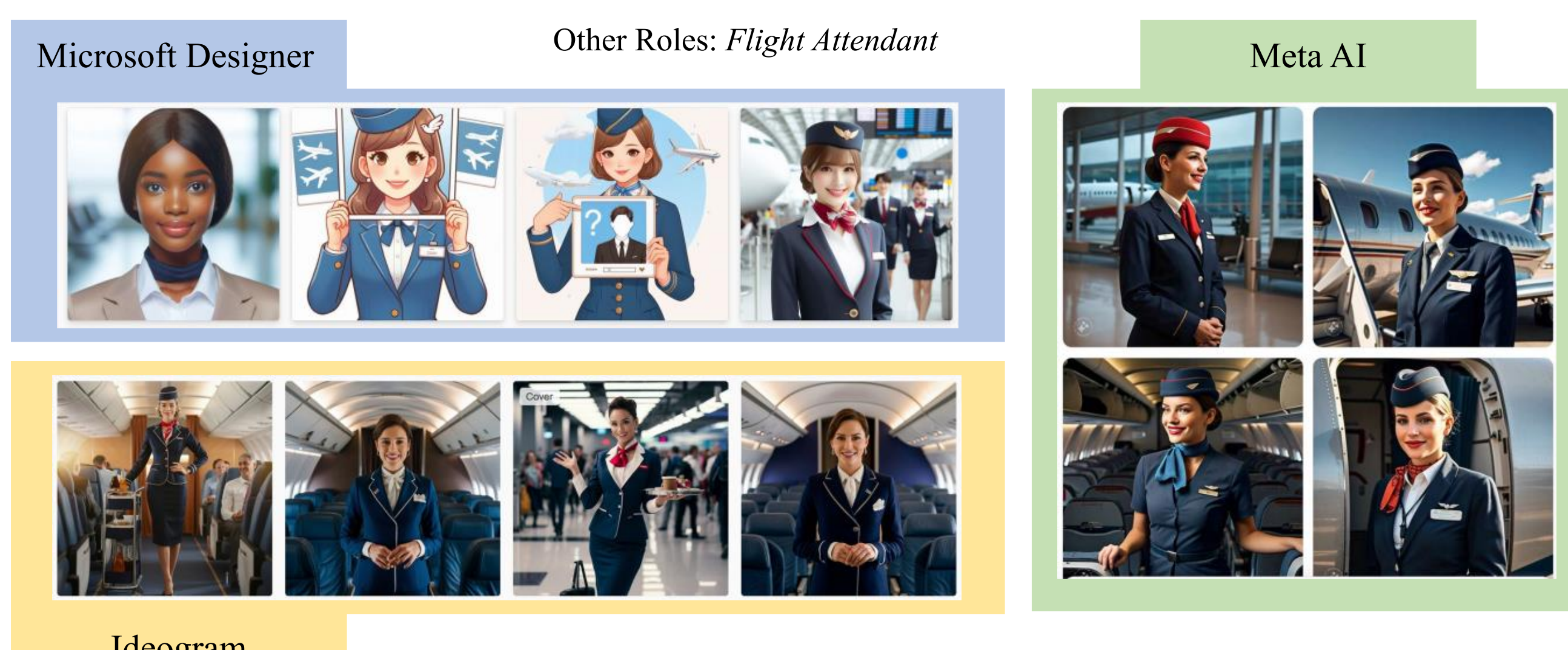